\documentclass[11pt]{article}

\usepackage[utf8]{inputenc}
\usepackage[T1]{fontenc}

\usepackage{arxiv}

\usepackage{amsmath}
\usepackage{amssymb}
\usepackage{bm}
\usepackage{graphicx}
\usepackage{booktabs}
\usepackage{array}
\usepackage{tabularx}
\usepackage{listings}
\usepackage{subcaption}
\usepackage{float}
\usepackage{enumitem}
\usepackage{natbib}

\usepackage{hyperref}
\usepackage{url}
\hypersetup{
  colorlinks=true,
  linkcolor=arxivaccent,
  citecolor=arxivaccent,
  urlcolor=arxivaccent,
  filecolor=arxivaccent,
  breaklinks=true,
}

\definecolor{codebg}{rgb}{0.95,0.95,0.97}
\definecolor{codegray}{rgb}{0.5,0.5,0.5}
\lstdefinestyle{transcript}{
    backgroundcolor=\color{codebg},
    basicstyle=\ttfamily\footnotesize,
    breaklines=true,
    captionpos=b,
    numbers=none,
    frame=single,
    rulecolor=\color{codegray},
    tabsize=2,
}
\newcommand{\attmem}{\textsc{AttMem}}
\newcommand{\patha}{\textsc{Path~A}}

\title{Parametric Multimodal User Memory:\\ Storing What Captions Cannot Carry}

\author{%
  \begin{tabular}{@{}c@{\hspace{4em}}c@{}}
    Bojie Li & Noah Shi \\
    Pine AI & University of Washington
  \end{tabular}%
}
\date{}
\runningtitle{Parametric Multimodal User Memory}

\begin{document}
\maketitle

\begin{abstract}
A personalized agent needs a \emph{user memory}: a persistent model of who its user is. Today it is almost always \emph{text} --- transcripts and captions retrieved by similarity. This serves the \emph{captionable} half of a person (``my cat is named Bibi''), but discards the \emph{perceptual} half no caption can hold: how a voice sounds, how a face reads across age and lighting, how tired someone sounds. We measure this loss across five modalities: a strong caption-based re-identifier recovers as little as $0.11$ of a dedicated encoder's recall, collapsing toward chance on non-nameable signals.

We instead \textbf{ground} perceptual memory in the model, decomposing recall into two subproblems: a vision-language model grounds the referent in context (\emph{what} and \emph{where}), and a dedicated encoder extracts an identity \emph{key} (\emph{who}), stored as one inline token read by attention at generation with no external round-trip. Neither suffices alone --- the VLM identifies cross-age faces at only $0.54$ recall where a face encoder reaches $0.81$, and an ungrounded encoder recognizes a two-person-scene referent at $0.05$ --- yet together they reach correct-region oracle ($0.96$), generalizing to multi-speaker audio and video. The recognition core is \emph{training-free}: it reproduces the encoder's recall on any frozen model at $\mathcal{O}(1)$ registration cost. On \textbf{PerceptMem} ($12$ domains, $1{,}080$ tasks) perceptual identity is capacity-limited (recall $\approx\min(1,k/M)$ of the encoder's ceiling) while exact facts are binding-limited: identity belongs in a parametric bank, facts in a text store. The two memories compose cleanly: an agent with both can remember not only what its user said, but also what they are like.
\end{abstract}

\begin{center}
\small
Code: \url{https://github.com/19PINE-AI/multimodal-user-memory} \\[2pt]
Website: \url{https://01.me/research/multimodal-user-memory}
\end{center}

\section{Introduction}
\label{sec:intro}

\begin{figure}[t]
  \setlength{\abovecaptionskip}{3pt}
  \centering
  \includegraphics[width=0.98\linewidth]{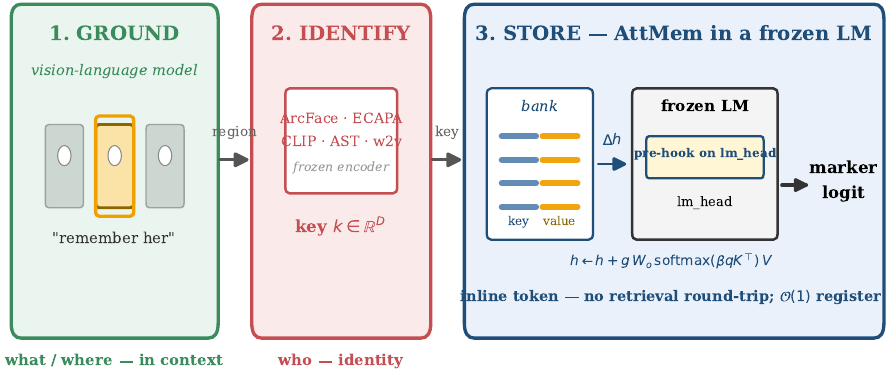}
  \caption{A vision-language model grounds the referent the user means in a cluttered scene (\emph{what}/\emph{where}); a purpose-built encoder turns the grounded region into an identity key (\emph{who}); and \attmem{} stores the key on a frozen language model, reading it back as a single inline token. Neither component suffices alone --- the VLM is a weak identity encoder, and a bare encoder cannot pick the referent out of a scene; grounded, they recover the correct-region oracle (Figure~\ref{fig:blindspots}). The mechanism is detailed in Figure~\ref{fig:arch}.}
  \label{fig:overview}
\end{figure}

When an AI agent ``remembers'' something about the person it serves, what does it actually keep? In almost every system today, the answer is \emph{text}. The dominant conversational memory architectures---Mem0~\citep{mem0}, MemoryLLM~\citep{memllm}, MemGPT / Letta~\citep{memgpt,letta}, M3-Agent~\citep{m3agent}, and related work surveyed in LongMemEval~\citep{longmemeval}---convert what the agent sees or hears into transcripts (via ASR~\citep{whisper}) and captions (via models such as BLIP-2~\citep{blip2} or LLaVA~\citep{llava}). These fragments are stored in a sentence-encoder vector index~\citep{sbert} and retrieved later by similarity. Similarly, personalized vision-language systems (MyVLM~\citep{myvlm}, Yo'LLaVA~\citep{yollava}, MC-LLaVA~\citep{mcllava}, Online-PVLM~\citep{onlinepvlm}, RAP~\citep{rap}) also rely on a textual backbone: a \emph{named} concept (e.g., ``Bibi'') serves as the handle, while the underlying perception remains an afterthought.

This strategy works for the \emph{captionable} part of a person --- information like ``my cat is named Bibi,'' ``I am vegetarian'' pass through text intact. However, it is not enough for the rest: how a voice sounds or how a face reads across age and lighting cannot be captured in text alone. The description ``a brown-haired man'' cannot tell two brown-haired men apart. Furthermore, a re-identification note from a strong captioner recovers a perceptual identity at as little as $0.11$ of a purpose-built encoder's recall (Section~\ref{sec:results})---the loss is a property of the channel, not the wording. This \emph{perceptual} half is disregarded in current systems: it is flattened into a caption and lost, or is handed to a recognizer that returns a bare label like ``speaker 47'' which the model never sees. The missing piece is a memory that stores a perception \emph{as a perception}.

\paragraph{Our proposal: grounded, parametric, multimodal user memory.} We keep each perception in its native modality and decompose remembering into three steps (Figure~\ref{fig:overview}): a vision-language model \emph{grounds} the user's intended referent in context (\emph{what} and \emph{where}). \textbf{Grounding is the act of isolating the referent before encoding it.} A dedicated encoder (ArcFace~\citep{arcface}, ECAPA-TDNN~\citep{ecapa}, CLIP~\citep{clip}) turns the grounded region into an identity \emph{key} (\emph{who}); and \attmem{}, a small per-modality memory bank on a frozen language model, \emph{stores} the key and reads it back by attention at the output head as a single marker token. Recall is then a token the model conditions on --- produced inside generation with no caption and no retrieval round-trip, registered in one $\mathcal{O}(1)$ tensor append with no per-user training. \attmem{} borrows the attention-over-a-bank shape of $k$NN-LM~\citep{khandelwal2020nearest} and Memorizing Transformers~\citep{wu2022memorizing}, aimed at a new target: a user's persistent perceptual identity.

\begin{figure}[t]
  \centering
  \includegraphics[width=0.86\linewidth]{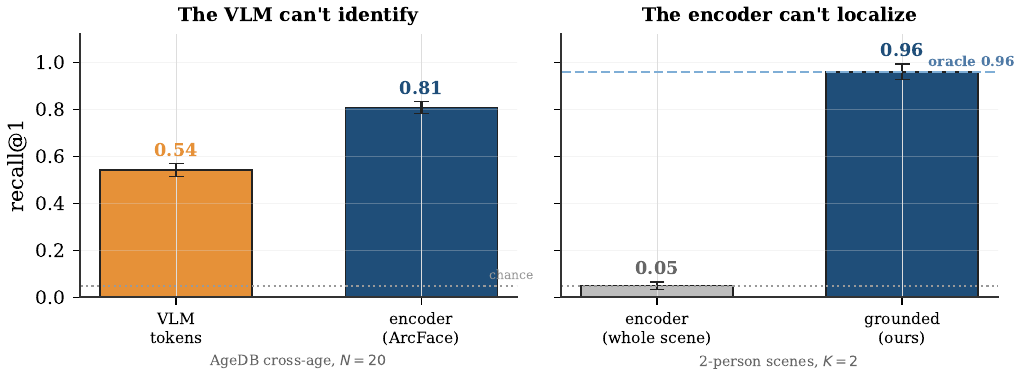}
  \caption{\textbf{Neither component works alone; grounding covers both blind spots.} Two targeted experiments isolating the failures of each component. \emph{Left (identity):} on AgeDB cross-age faces ($N{=}20$), the VLM's vision tokens are a weak identity encoder (recall $0.54$) while a dedicated face encoder achieves $0.81$ recall. \emph{Right (localization):} in two-person scenes ($K{=}2$), the performance of a bare encoder embedding the whole scene is close to random ($0.05$), while grounding the referent first achieves near-perfect performance ($0.96$). The division of labour is essential: the VLM excels at localization but struggles with identity invariance, while the specialist encoder is highly invariant but cannot locate the target. Detailed ablations appear in Figure~\ref{fig:agentic}.}
  \label{fig:blindspots}
\end{figure}

\paragraph{Why grounding is necessary.} A perceptual memory cannot simply encode whatever the agent observes, because a raw perception mixes the identity worth remembering with the context around it. The same friend photographed at a coffee shop and while surfing have almost no pixels in common, so a whole-image embedding would place them far apart. However, we recognise the person to be the same at a glance by \emph{focusing on the face} and discarding the background. The same principle applies to audio: a voice on a phone call and one in a podcast differ substantially in terms of words and acoustics, yet a listener reliably identifies the speaker. This creates a division of labour (Figure~\ref{fig:blindspots}): a vision-language model identifies \emph{which} region to attend to but is a weak identity encoder, while a specialist encoder produces condition-invariant representations but cannot locate the referent in a cluttered scene. Grounding lets each component cover the other's blind spot, achieving recognition that neither can reach alone.

\begin{figure}[t]
  \centering
  \includegraphics[width=0.99\linewidth]{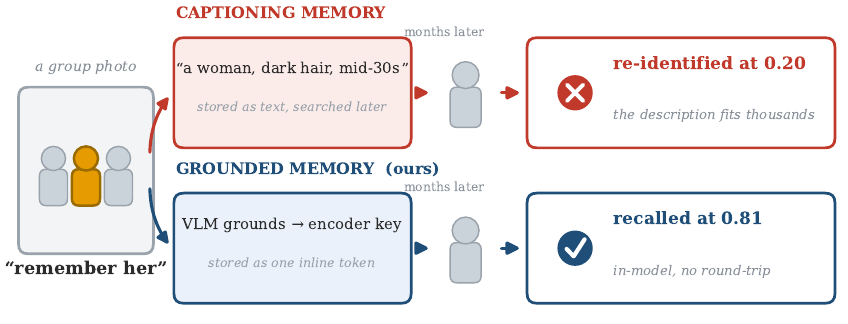}
  \caption{\textbf{A concrete example.} Given the instruction ``Remember her'' in a group photo, a captioning memory stores a generic description that matches thousands of individuals, leading to near-random re-identification performance. Our grounded memory instead stores a single inline token that supports accurate recall of the correct person across different conditions. The numbers shown are paired re-identification accuracies (Section~\ref{sec:results}).}
  \label{fig:worked}
\end{figure}

Figure~\ref{fig:worked} contrasts the two methods on one ``remember her'' task.

\paragraph{Contributions.}
\begin{itemize}[leftmargin=*,itemsep=2pt,topsep=2pt]
\item \textbf{Perceptual user memory must be grounded, not captioned.} We argue and empirically demonstrate that text is fundamentally insufficient for perceptual user content. Effective recall instead requires a decomposition into VLM grounding, specialist encoding, and in-model storage (Sections~\ref{sec:intro}--\ref{sec:method}).
\item \textbf{A grounded memory that recovers oracle recognition in context.} By combining a VLM for referent grounding, a specialist encoder for identity extraction, and an inline attention token for storage, our method achieves near-oracle accuracy in cluttered scenes---where whole-scene encoding is near chance. This holds across vision, audio, and video (Section~\ref{sec:results}).
\item \textbf{A training-free, $\mathcal{O}(1)$ in-model recognition core.} Our memory is read via attention at the output head with \emph{zero} gradient updates. It reproduces the specialist encoder's performance across ten frozen model families, with constant-time insertion and recall. This is far cheaper than keeping raw perceptions in context (Sections~\ref{sec:method}--\ref{sec:results}).
\item \textbf{PerceptMem benchmark and a capacity-law router.} We introduce PerceptMem, a benchmark spanning $12$ perceptual domains and $1{,}080$ tasks across five modalities. The benchmark includes text-only and random chance baselines. In addition, we propose two capacity laws that route perceptual identity to the parametric bank and factual knowledge to a text store (Sections~\ref{sec:bench},~\ref{sec:discussion}).
\end{itemize}

\section{Decomposing perceptual recall: grounding and identification}
\label{sec:why}

Why does the text-based route fail so consistently? The problem is not lack of effort but a fundamental structural limitation. Consider the most challenging case for voice identification: the agent has ten enrolled voice samples on file, and a new recording arrives from a different room, on a different day, with different acoustics. It must determine which of the ten speakers it belongs to. An agent has four options (Table~\ref{tab:routes}). Three of them discard the raw perceptual signal that distinguishes one voice from another. Only the fourth route preserves this signal --- yet doing so immediately reveals a deeper issue: every approach implicitly conflates two distinct sub-problems. Recognizing a perception requires both (1) localizing the referent in a cluttered context and (2) encoding its identity. No single component performs both tasks well.

\begin{table}[h]
\footnotesize\centering
\caption{Four routes for remembering a perception, plus ours (citations in text). Only routes that keep the raw perception can discriminate across conditions. Only embedding retrieval and our method inherit the encoder's recall; ours additionally localizes the referent with a VLM and reads the identity back \emph{inside} the model as a token the model conditions on, which is training-free and in $\mathcal{O}(1)$.}
\label{tab:routes}
\setlength{\tabcolsep}{5pt}
\begin{tabular}{@{}llccc@{}}
\toprule
\textbf{Route} & \textbf{What it stores} & \textbf{Keeps percept?} & \textbf{Cross-condition} & \textbf{Reg.\ cost} \\
\midrule
Caption-and-search    & text description        & no  & chance        & $\mathcal{O}(1)$ \\
Recognize-and-label   & a bare label            & no  & label only    & $\mathcal{O}(1)$ \\
Per-concept tuning    & tuned token/adapter     & yes & strong        & gradient / identity \\
Embedding retrieval   & raw embedding           & yes & encoder-limited & $\mathcal{O}(1)$ \\
\textbf{Parametric (ours)} & embedding $+$ LM value & yes & \textbf{$=$ encoder, in-model} & $\mathcal{O}(1)$ \\
\bottomrule
\end{tabular}
\end{table}

\textbf{The three routes that discard the signal.} \emph{Caption-and-search}~\citep{mem0,memllm} writes the voice down (``a male voice, mid-30s, slight Eastern-European accent'') and later retrieves by similarity over the words. While the description is true, it is far too generic: it matches thousands of speakers, so cosine similarity over captions cannot reliably distinguish even ten users. \emph{Recognize-and-label}~\citep{m3agent} calls a recognizer (ArcFace~\citep{arcface}, ECAPA-TDNN~\citep{ecapa}) and stores the label it returns (``speaker 47''). The model can recall the assigned label, but it never stores the actual perception so it cannot compare a new voice against previous ones. \emph{Per-concept tuning}~\citep{yollava,myvlm} fine-tunes a dedicated token or adapter per concept. While effective, this approach is impractical. Each identity requires gradient updates, and the resulting artifact is concept-specific. This cost profile does not scale for a user who registers dozens of new people, voices, and objects.

\textbf{The one that keeps the signal, and what it still lacks.} \emph{Embedding retrieval} bypasses text entirely: it indexes by the raw perceptual embeddings, and retrieves by cosine similarity. This is the only text-free route that preserves the original perception, and serves as our strongest baseline. It succeeds to the degree that the encoder is invariant across recording conditions, and fails when the invariant is broken. On a 2180-identity face pool, it reaches 0.95 recall at ten identities, but performance degrades as the pool grows. Crucially, this method assumes the perception has \emph{already} been cleanly isolated---i.e., it receives a clean voiceprint or a tight face crop. In realistic cluttered scenes, this isolation step itself is extremely difficult. A bare encoder is helpless here: embedding an entire two-person photo yields only $0.05$ accuracy at identifying the intended person (Section~\ref{sec:results}). This is the second problem that all four routes conflate: \emph{localizing} the referent is not the same as \emph{identifying} it. Even the strongest text-free route only solves the latter.

Our approach builds on this identity step while addressing its two key shortcomings. We place a VLM in front to perform grounding (resolving and locating the intended referent), and we give the resulting identity a new home inside the model. Instead of retrieving a neighbor from an external index, we store the encoder key together with a dedicated marker token and read it back via attention directly at the model's output head (Section~\ref{sec:method}). The recalled identity thus becomes a native token the model can condition on during generation. Recognition performance itself remains unchanged --- our sharp read closely approximates a hard argmax over encoder cosine similarities --- but the memory now additionally provides grounding, an in-model representation, $\mathcal{O}(1)$ registration cost, and seamless portability across frozen models.

\section{Grounding perceptions in the model's own representation}
\label{sec:method}

We decompose perceptual recall into three steps, which we address in turn: (1) \emph{ground} the referent in its visual or auditory context using a VLM, (2) \emph{identify} it with a dedicated encoder, and (3) \emph{store} the resulting identity as a token that can be read directly inside a frozen language model. The first two steps leverage off-the-shelf components placed at their points of strength. The third step---\attmem{}, our core contribution---is the novel mechanism introduced in this paper.

\paragraph{Grounding: resolving the referent in context.} Perceptions rarely arrive pre-cropped. When a user points to one person in a group photo, refers to the painting on the left, or mentions ``the voice that just spoke,'' the system must first resolve this context-dependent reference into a specific region before encoding can begin. This is precisely what vision-language models excel at, and what specialist encoders cannot do: the VLM grounds the user's referent against the surrounding scene and the user's words, returning a localized region (a bounding box for images, or a temporal span for audio). We use the VLM's native grounding capability for this step. Where available, we further refine the region using a dedicated detector (e.g., RetinaFace~\citep{retinaface} for faces, from the same InsightFace suite as ArcFace). This handles bounding box imprecision and unaligned scenes before identity encoding. The necessity of this grounding step is strongly supported by experiments: encoding an entire ungrounded scene recognizes the intended referent at only $0.05$ recall for faces and $0.11$ for paintings---close to random---because a single embedding cannot disambiguate the chosen referent from distractors (Section~\ref{sec:results}).

\paragraph{Identification: a dedicated encoder, not the model's own tokens.} Once the referent has been grounded to a specific region, we extract its identity using a dedicated perceptual encoder. The resulting L2-normalized embedding serves as the identity \emph{key}. We use ArcFace for faces, ECAPA-TDNN for speakers, CLIP for painting style, AST~\citep{ast} for acoustic scenes, and a wav2vec2 emotion encoder~\citep{wav2vec2} for tone. It is tempting to skip this step and instead extract identity directly from the VLM's own context-conditioned representations (i.e., asking the model to ``output the embedding of the face''). However, this does not work well in practice. A linear probe on the VLM's hidden states achieves only $0.25$ recall at identifying the correct referent (raw hidden state $0.14$), far below the oracle performance of $0.80$. Similarly, the VLM's native vision tokens achieve only $0.54$ recall on AgeDB cross-age faces, compared to $0.81$ for ArcFace at the same scale $N{=}20$ (Figures~\ref{fig:blindspots},~\ref{fig:crossdomain}). This performance gap persists even in end-to-end tests inside a full VLM (\S\ref{sec:when}). In short, VLMs are strong localizers but weak identity encoders. Strong identity extraction therefore requires a dedicated perceptual encoder kept separate from the model's internal representations.

\begin{figure}[t]
  \centering
  \includegraphics[width=0.99\linewidth]{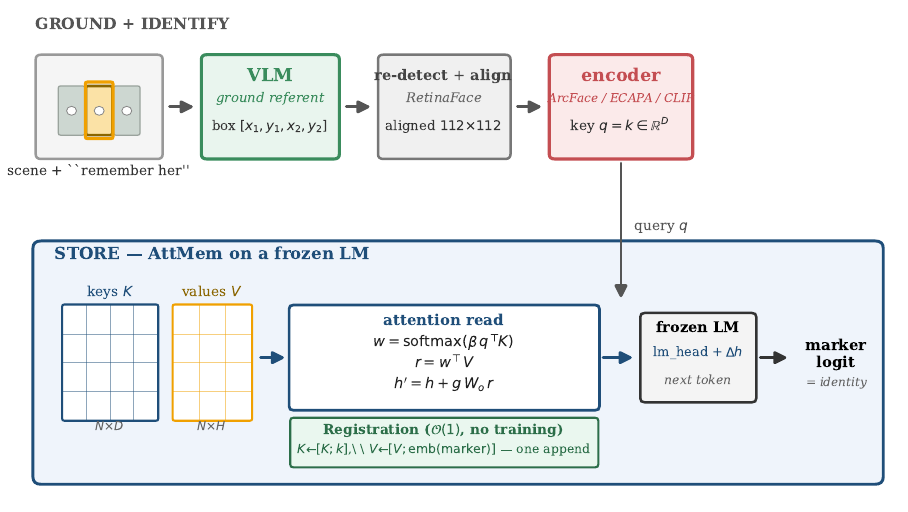}
  \caption{The grounded memory in detail. \textbf{Ground + identify (top):} the VLM resolves the referent to a box, a modality detector re-detects and aligns the region (RetinaFace for faces), and a frozen perceptual encoder (ArcFace / ECAPA / CLIP) maps it to an L2-normalized key $q{=}k\in\mathbb{R}^{D}$. \textbf{Store (bottom):} \attmem{} holds keys $K$ ($N{\times}D$) and values $V$ ($N{\times}H$, the model's own marker-token embeddings); the attention read $\bm{w}{=}\mathrm{softmax}(\beta\,\bm{q}^{\!\top}\!K)$, $\bm{r}{=}\bm{w}^{\!\top}V$, $\bm{h}'{=}\bm{h}{+}g\,W_o\bm{r}$ adds a residual at the output head that biases the next token toward the matching marker. The read has no trainable parameters ($\beta,g$ are constants); registration is a single $\mathcal{O}(1)$ append.}
  \label{fig:arch}
\end{figure}

\paragraph{Storage: an in-model attention memory.} The identity key is stored and retrieved using \attmem{}, a table of key-value rows on a frozen language model (Figure~\ref{fig:arch}). Each row $i$ contains a \emph{key} $\bm{k}_i$ (the encoder embedding above) and a \emph{value} $\bm{v}_i$ (the language model's own embedding for a marker token assigned to that identity). Storing the model's native vector as the value (rather than an arbitrary learned one) is a deliberate choice. It makes the eventual logit boost for the right marker clean and self-reinforcing, instead of relying on an accidental alignment of two unrelated vectors.

Reading the memory is one step of attention. When a localized perception arrives at generation time, right before the output head, the model forms a query $\bm{q}$ from its encoder key and computes
\[
\bm{w} = \mathrm{softmax}\!\bigl(\beta\,\bm{q}^\top\bm{K}\bigr), \qquad
\bm{r} = \bm{w}^\top\bm{V}, \qquad
\bm{h}' = \bm{h} + g\cdot W_o\,\bm{r},
\]
where $\bm{K},\bm{V}$ stack the bank's keys and values, $\beta$ is the attention sharpness, $g$ a residual gain, and $W_o$ an optional output projection --- the identity in all our experiments, kept in the equation only to mark where a projection would sit. In words, the perception softly looks up the bank, pulls back a blend of the matching markers' model-side vectors, and nudges the next-token prediction toward them. The blend is the entire read, and it adds \emph{no trainable parameters}: $\beta$ and $g$ are two constants, and the values are the model's own marker embeddings.

\paragraph{Design choices for a faithful read.} Four design decisions determine whether the read recovers the encoder's nearest neighbour cleanly with no training. 
\begin{enumerate}
  \item \emph{Do not} divide the attention logits by $\sqrt{D}$: with L2-normalized keys, that flattens every cosine similarity into near-uniform attention.
  \item Use a high sharpness $\beta$: this makes the attention a near-hard argmax over the bank rather than a diffuse average.
  \item Attach the hook at the output head (not at an intermediate layer), so the residual reaches the logits undiluted.
  \item Set a large gain $g$ so the retrieved marker dominates the hidden state, producing an exact read. On models with untied output embeddings (whose output-embedding norms run smaller), an even larger $g$ is required (Appendix~\ref{app:tables}, Table~\ref{tab:gainrobust}).
\end{enumerate}
With these four choices, the read reproduces the encoder's recall exactly (\S\ref{sub:core}), and registering a user reduces to a single row append with no training required.

\begin{figure}[t]
  \centering
  \includegraphics[width=0.62\linewidth]{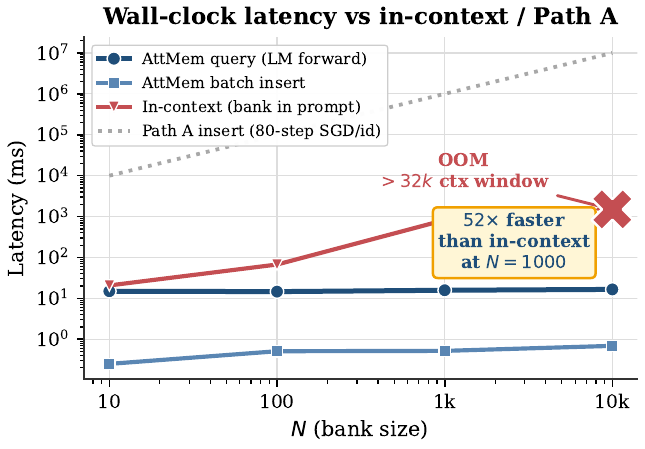}
  \caption{Recall latency remains near constant ($\sim$15\,ms) as bank size increases, as it is dominated by the frozen model's own forward pass. Insertion is $\mathcal{O}(1)$. Feeding the same registered perceptions in as context grows linearly and exhausts the $32$k context window (Qwen2.5-3B) at ten thousand identities. Per-concept fine-tuning (\patha{}-style) costs gradient steps per identity. Log--log axes.}
  \label{fig:latency}
\end{figure}

\paragraph{Registration versus fine-tuning.} The cost of writing to memory is a critical practical consideration (Figure~\ref{fig:latency}). Adding an identity is a single \texttt{torch.cat} operation (about half a millisecond to add a thousand), and recall is constant-time regardless of bank size, as the lookup is a small matrix multiply that is dwarfed by the model's own forward pass ($\sim$15\,ms either way). By contrast, the natural alternative of feeding all previously registered perceptions into the model as context grows linearly with the number of identities per query and quickly exhausts the context window. Our mechanism is 52$\times$ faster at a thousand identities, and is the only one that remains usable at ten thousand. A memory system intended for a companion agent---which must be updated after every conversation---must be inexpensive to write to.

\section{PerceptMem: a cross-condition perceptual-memory benchmark}
\label{sec:bench}

To measure perceptual recall we introduce \textbf{PerceptMem}, a benchmark spanning five sub-modalities. Each sub-modality was chosen because the discriminative signal is known to be severely degraded or destroyed when converted to text captions (Table~\ref{tab:datasets}). The benchmark covers both vision and audio, as well as identity and style, spanning physical appearance and affective cues:

\begin{table}[h]
\small\centering
\caption{The five PerceptMem sub-modalities. Each is included because no caption can carry its discriminating signal.}
\label{tab:datasets}
\begin{tabularx}{\linewidth}{@{}llX@{}}
\toprule
\textbf{Sub-modality} & \textbf{Source / encoder} & \textbf{Why text cannot carry it} \\
\midrule
Face across age \& lighting & LFW+AgeDB / ArcFace & ``brown-haired man'' fits thousands \\
Painter style & WikiArt / CLIP-mid & no caption separates early vs.\ late Monet \\
Speaker across recordings & LibriSpeech / ECAPA-TDNN & timbre is not transcribable \\
Acoustic scene & ESC-50 / AST & ``traffic noise'' fits every street \\
Tone of voice & RAVDESS / wav2vec2 & ``sounded tired'' lacks a personal baseline \\
\bottomrule
\end{tabularx}
\end{table}

The benchmark uses standard, widely adopted datasets and encoders: LFW~\citep{lfw} and AgeDB~\citep{agedb} read with ArcFace~\citep{arcface}, WikiArt~\citep{wikiart} with mid-layer CLIP~\citep{clip}, LibriSpeech~\citep{librispeech} with ECAPA-TDNN~\citep{ecapa}, ESC-50~\citep{esc50} with AST~\citep{ast}, and RAVDESS~\citep{ravdess} with a wav2vec2 emotion encoder~\citep{wav2vec2}.

PerceptMem is deliberately designed to evaluate \emph{memory} rather than raw perception. Every sub-modality provides a fixed, off-the-shelf encoder that all methods can use as black-box infrastructure, ensuring that no approach gains an advantage from using a stronger encoder. The only allowed operations are \texttt{register} and \texttt{recall}. All data is cross-session: each registered sample and its corresponding query come from different recordings. Training identities never overlap with evaluation identities. For a memory of size $N$, we register one sample per identity and then issue cross-condition queries, asking the method to retrieve the correct identity. Throughout this paper, \emph{recall} refers to recall@1---the fraction of queries for which the top-scoring registered identity is correct. For retrieval-based methods this is the encoder's nearest neighbor; for our in-model mechanism it is the argmax over the registered marker token logits. For tone of voice, each registered ``identity'' is a (speaker, emotional state) pair. Successful recall therefore requires matching a paralinguistic state across different utterances---a capability that per-utterance captions cannot achieve without access to the user's personal history.

For statistical rigor we expand the five conceptual sub-modalities into a $12$-domain, $1{,}080$-task benchmark across multiple datasets and encoders. Each domain is defined as a (dataset, encoder) pair; their pool sizes and per-domain recall statistics are provided in Appendix Table~\ref{tab:domains} (see also Section~\ref{sec:results} and Figure~\ref{fig:crossdomain}). We report a strong text-only baseline and a random guessing floor on every comparison. Our main reference point throughout is \textbf{embedding retrieval}---i.e., using the same encoder with cosine nearest-neighbor search over the registered keys. We refer to this as the \emph{encoder ceiling}, since it represents the best possible performance achievable using the encoder's own similarity metric. Our in-model read is designed to reproduce this ceiling rather than surpass it.

\section{Empirical evaluation}
\label{sec:results}

We structure our evaluation around the central claims of the grounded memory approach. First, we show that the performance drop from using text is both real and substantial: captions are fundamentally unable to carry perceptual identity (\S\ref{sub:textcant}). Second, we demonstrate that grounding effectively recovers this signal. Combining VLM-based grounding with a dedicated encoder reaches the correct-region oracle in cluttered scenes (\S\ref{sub:factored}), and the same decomposition generalizes successfully to audio and video (\S\ref{sub:modalities}). Third, we establish that the recognition core is the dedicated encoder itself, and that faithfully reading it in-model is a deliberate design strength, not a limitation (\S\ref{sub:core}). Matching the encoder's performance is the goal: it shows we have preserved the encoder's quality while adding the benefits of grounding, in-model access, training-free registration, and portability across frozen models. We then show that the in-model store enables seamless composition with text memory (\S\ref{sub:compose}), matches the strongest pipeline baselines while significantly outperforming in-context image methods on both accuracy and cost (\S\ref{sub:cost}), and fully preserves the model's original text behavior (\S\ref{sub:nonreg}). Finally, we integrate all components into an end-to-end multi-session agent (\S\ref{sub:agent}).

\subsection{Text captions cannot carry perceptual identity}
\label{sub:textcant}

\begin{figure}[t]
  \centering
  \includegraphics[width=0.86\linewidth]{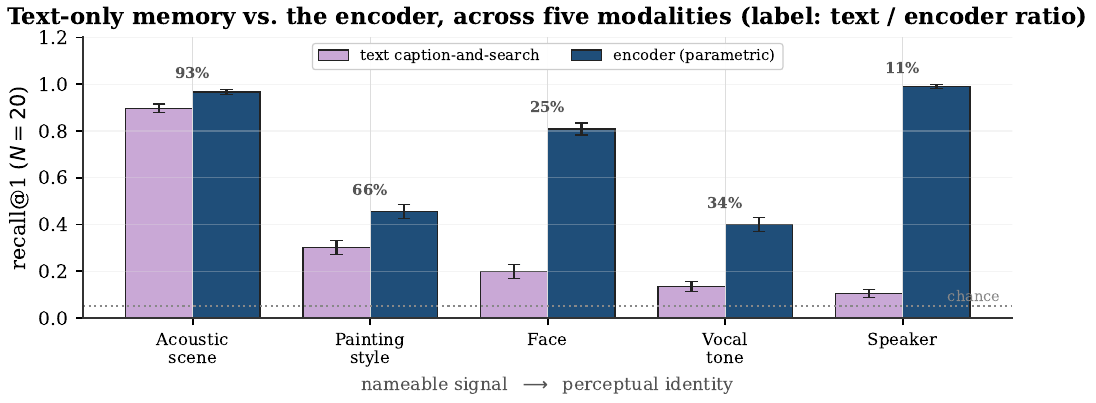}
  \caption{Text-only caption-and-search versus the parametric encoder, paired on identical draws across all five modalities ($N{=}20$, $30$ draws, $95\%$ CI); the label above each pair is the text/encoder ratio. Modalities are ordered by how nameable the signal is. Text nearly matches the encoder on the nameable acoustic scene ($93\%$) and degrades toward random chance on perceptual identity (speaker $11\%$), tracing the router's boundary between the text store and the parametric bank.}
  \label{fig:textablation}
\end{figure}

A common baseline is to describe the perception in words, store the resulting text, and retrieve by semantic similarity. We evaluate this approach across \emph{all five modalities}, giving it the strongest possible setup: an LLM (Qwen2.5-VL~\citep{qwen25vl} for images, Qwen2.5-Omni~\citep{qwen25omni} for audio) generates a re-identification caption, which is then embedded with a sentence encoder. Recognition is performed via cosine nearest-neighbor search over these captions and compared directly against the dedicated encoder on the same samples. The results reveal a clear pattern (Figure~\ref{fig:textablation}). Caption-based recall approaches that of the dedicated encoder when the signal is easily \emph{nameable} (e.g., an acoustic scene described as ``someone typing on a keyboard''). However, performance drops sharply toward chance when the signal is a fine-grained \emph{perceptual identity} (e.g., a specific voice, where descriptions such as ``high-pitched with an accent'' apply to many speakers). In short, text is a competent memory for concepts that can be described verbally, but a poor one for perceptual identity that cannot. This is exactly the distinction our capacity-law router exploits when deciding between the text store and the parametric memory bank (\S\ref{sec:discussion}).

\subsection{Grounding recovers oracle recognition in cluttered scenes}
\label{sub:factored}

\begin{figure}[t]
  \centering
  \includegraphics[width=0.82\linewidth]{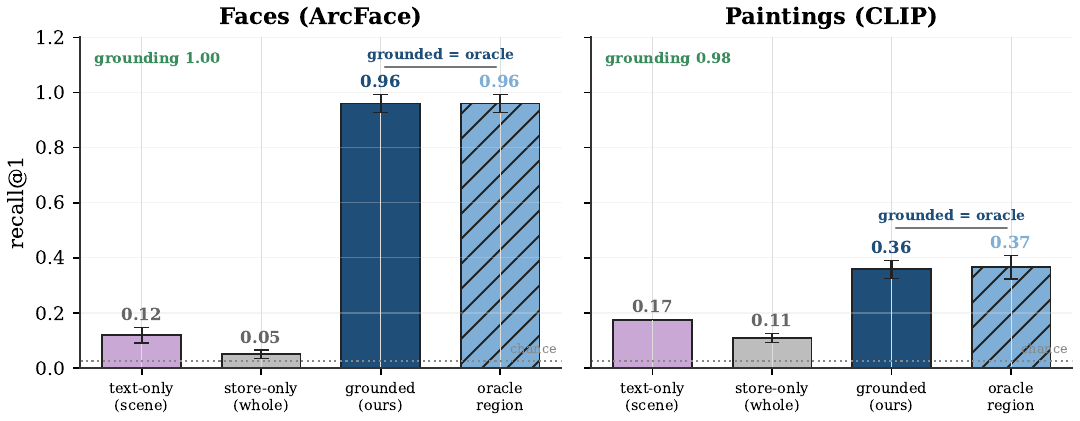}
  \caption{Ablation of the grounded memory on cluttered scenes ($M{=}40$, two referents per scene). Performance increases left to right: \emph{text-only}, where the VLM describes the referent from the whole scene and the captions are matched; \emph{store-only}, where the whole scene is embedded with no grounding; \emph{grounded}, our ground $+$ identify $+$ store strategy; and the correct-region \emph{oracle}. Both text-only and store-only perform near random guessing, while grounding reaches the oracle-level performance. Retrieval uses the encoder's nearest neighbor in all cases, so the large gain comes from grounding itself, not the matching method. Faces: ArcFace with landmark alignment, $5$ seeds; paintings: CLIP with no alignment, $3$ seeds. Error bars show $\pm$95\% CI.}
  \label{fig:agentic}
\end{figure}

In deployment, perceptions rarely arrive pre-cropped. The referent usually appears among other people or objects, and the system must first resolve which one the user means before encoding can begin. An ablation study isolates the contribution of each component (Figure~\ref{fig:agentic}). A \emph{text-only} memory, in which the VLM captions the referent from the full scene and retrieval is performed by text similarity, achieves only $0.12$ recall on faces and $0.17$ on paintings. A \emph{store-only} memory that embeds the entire ungrounded scene performs even worse, close to random guessing, because a single embedding cannot disambiguate the intended referent among distractors. 

In contrast, \emph{grounding} the referent first and then identifying it with the encoder recovers the correct-region oracle performance in both domains: $0.96$ recall on faces and $0.36$ on paintings (each matching its respective oracle). The relatively low ceiling on paintings is due to CLIP's limitations, not grounding. The VLM successfully localizes the correct region on nearly every trial. Since retrieval is performed using the same encoder nearest-neighbor in both cases, the large performance gain comes entirely from grounding, not from the matching mechanism.

\begin{figure}[t]
  \centering
  \includegraphics[width=0.9\linewidth]{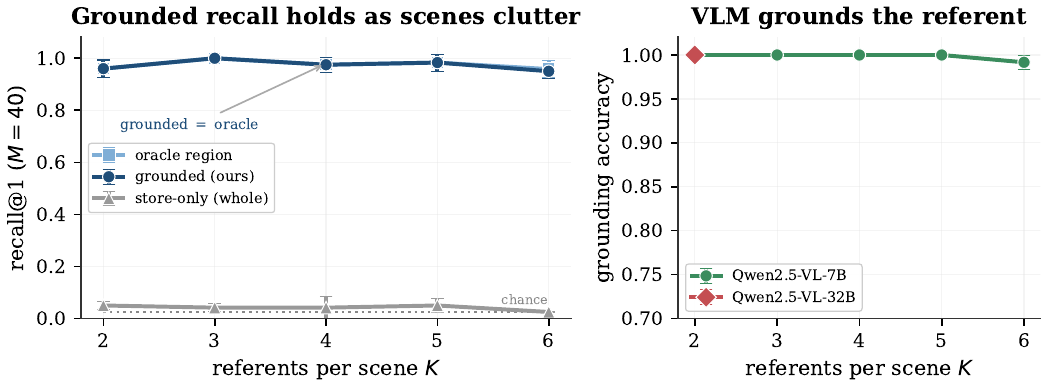}
  \caption{The grounded advantage is robust to scene clutter and VLM scale (faces, $M{=}40$). \textbf{Left:} As referents per scene grow from two to six, grounded recall (navy) closely tracks the correct-region oracle (light blue, nearly coincident), while whole-scene recall remains near the floor. \textbf{Right:} Grounding accuracy stays at $1.00$ through $K{=}5$ and only drops slightly to $0.99$ at $K{=}6$. Performance is identical for Qwen2.5-VL-7B and Qwen2.5-VL-32B, showing that the decomposition does not require a large grounding model.}
  \label{fig:density}
\end{figure}

The advantage of grounding is robust to increasing scene clutter and to different VLM scales (Figure~\ref{fig:density}). As the number of referents per scene grows from two to six, grounding accuracy remains perfect ($1.00$) up to $K{=}5$ and drops only marginally at $K{=}6$ ($0.99$). Grounded recall closely tracks the oracle ($0.95$--$1.00$) across all densities, while whole-scene embedding recall stays near the floor. In contrast, text-only recall degrades as clutter grows, falling from $0.12$ at $K{=}2$ to $0.05$ at $K{=}3$ as the caption must disambiguate more distractors. Grounding accuracy is already saturated with a $7$B VLM and unchanged at $32$B ($1.00$ in both cases; recall $0.960$ vs $0.958$). Overall, grounding is lossless and domain-general: the performance it enables is limited only by the quality of the downstream encoder---modest for CLIP-based style features and near-perfect for ArcFace-based identity.

\subsection{Grounding generalizes to audio and video}
\label{sub:modalities}

\begin{figure}[t]
  \centering
  \includegraphics[width=0.95\linewidth]{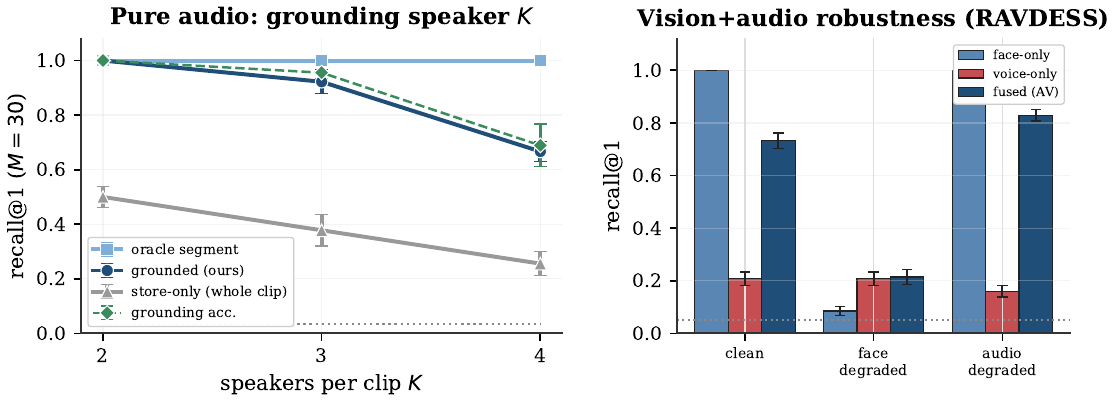}
  \caption{Grounding generalizes beyond vision. \textbf{Left --- pure audio:} An audio-LLM grounds the $K$-th speaker's time span in a multi-speaker VoxCeleb clip and ECAPA identifies it ($M{=}30$, 3 seeds). Grounded recall (navy) closely tracks the oracle segment at $K{=}2$--$3$ and degrades at $K{=}4$ as temporal grounding becomes difficult, while whole-clip encoding stays well below the grounded read. \textbf{Right --- vision+audio (RAVDESS video):} Each identity is represented by both a face and a voice key. Storing both modalities provides robustness to single-channel corruption: face-only recall collapses under distant/low-resolution views, voice-only performance is weak and further degrades under noise, but the fused memory consistently falls back to the surviving channel ($M{=}20$, 40 draws).}
  \label{fig:modalities}
\end{figure}

The proposed decomposition is not limited to vision. The same three steps apply when the referent is a \emph{speaker} in a multi-speaker audio recording or a \emph{person} in a video (Figure~\ref{fig:modalities}). For \textbf{pure audio}, we concatenate $K$ speakers from VoxCeleb~\citep{voxceleb} into a single clip and ask an audio-LLM (Qwen2.5-Omni) to localize the time span of the referenced speaker --- the direct analog of visual bounding-box grounding. We then identify the speaker using ECAPA. Grounded recall matches the oracle at $K{=}2$ ($1.00$) and remains strong at $K{=}3$ ($0.92$ recall, $0.96$ grounding accuracy). Performance drops at $K{=}4$ ($0.67$ recall, $0.69$ grounding accuracy) as temporal localization of the fourth speaker becomes difficult, while encoding the entire clip is far weaker (recall $0.50$ at $K{=}2$, falling to $0.26$ at $K{=}4$) and the correct-segment oracle scores $1.00$ throughout. This pattern holds in more realistic settings. On \emph{VoxConverse}~\citep{voxconverse}---in-the-wild YouTube conversations with natural turn-taking, overlaps, and background noise---grounding is harder (grounding accuracy $0.75$), but the grounded read still reaches $0.73$ recall, nearly doubling whole-window encoding ($0.38$) toward the oracle ($0.97$) across 60 conversations. The \emph{AMI} meeting corpus~\citep{ami} behaves the same way (grounded recall $0.75$ vs.\ whole-window recall $0.34$, grounding accuracy $0.88$). These results mirror the vision findings: grounding, rather than identity encoding, is the component most affected by increasing complexity.

For \textbf{video} (combined vision and audio), we use real recordings from RAVDESS, in which the same 24 actors are captured across eight different emotions. This provides genuine paired (face, voice) data per identity. We register each actor from one clip (using an ArcFace face key and an ECAPA voice key) and evaluate on a held-out clip of a different emotion. Face recognition is near-saturated ($1.00$), while cross-emotion voice recognition is challenging ($0.21$). As a result, simple fusion of the two modalities with equal weight does not outperform face recognition alone, and we do not claim that it does. The real value of storing \emph{both} modalities lies in robustness when one channel is corrupted---a capability that single-modality memory cannot provide (Figure~\ref{fig:modalities}, right). When the face is degraded to a distant, low-resolution view, face-only recall collapses to $0.09$, but the fused memory maintains performance by relying on the voice ($0.21$). Conversely, when audio is corrupted by noise, voice-only recall drops to $0.16$, but the fused memory falls back to the face ($0.83$). Fusion is the only representation that remains robust across degraded conditions---exactly the kind of real-world scenario a deployed companion agent must handle as people move off-camera or speak in noisy environments.

\subsection{The in-model read reproduces the encoder's recall}
\label{sub:core}

Once the referent is localized and encoded, the memory bank is responsible for recognition. With a sharp read, the attention closely approximates a hard argmax over the same encoder cosine similarities used by embedding retrieval. As a result, its recall matches that of the encoder---neither better nor worse. A paired evaluation protocol, using identical registrations and queries with randomized target slots, confirms this equivalence across bank sizes from $N{=}5$ to $N{=}1{,}000$ ($|\Delta|\le 0.001$ at each $N$, $20$ draws; Table~\ref{tab:headline}, Figure~\ref{fig:scaling}). The in-model memory ties embedding retrieval on both random banks and adversarial banks of near-identical look-alikes. A learned similarity metric trained on the same features performs similarly (Appendix~\ref{app:tables}). This tight match depends on using a sharp read. When the read is under-sharpened, performance begins to lag behind cosine similarity at $N\ge300$ (Appendix~\ref{app:tables}). The design choices described in Section~\ref{sec:method} eliminate this gap.

\begin{figure}[t]
  \centering
  \includegraphics[width=0.9\linewidth]{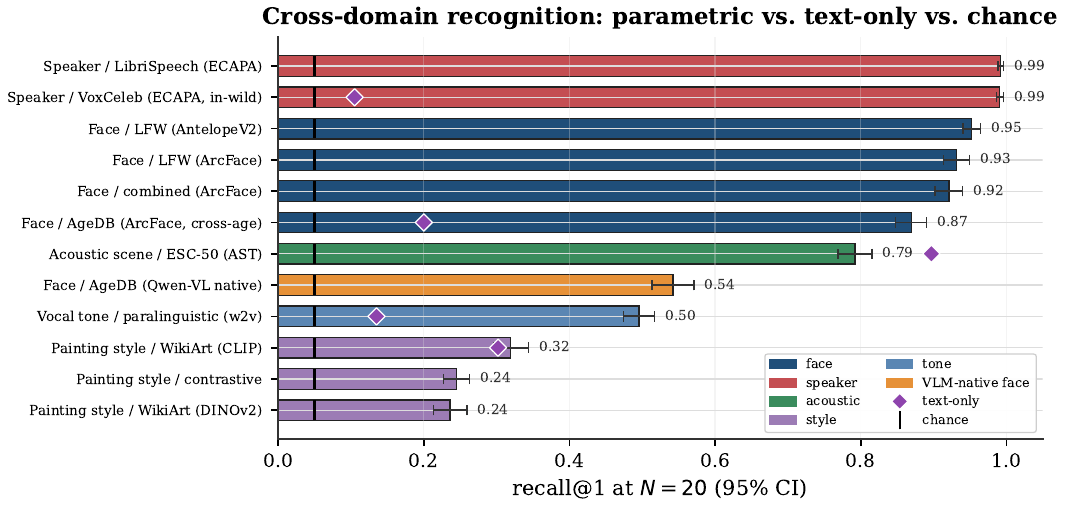}
  \caption{Recognition performance across $12$ perceptual domains and $5$ modalities ($1{,}080$ tasks; recall@1 at $N{=}20$, $30$ draws, $95\%$ CI). Results are colored by modality and compared against random chance ($1/N$, black ticks) and the text-only caption baseline where available (purple diamonds). The parametric memory closely tracks the dedicated encoder and substantially exceeds chance in all domains, far outperforming text captions on perceptual identity tasks (speaker, face, tone). The only inversion occurs for nameable acoustic scenes, where a text caption (e.g., ``typing on a keyboard'') slightly outperforms the weaker AST encoder --- precisely the boundary used by our router (\S\ref{sec:discussion}). Text works well for nameable categories, while the parametric bank is essential for fine-grained identity. For reference, the VLM's own native vision tokens (orange) perform substantially worse than a purpose-built encoder on AgeDB faces.}
  \label{fig:crossdomain}
\end{figure}

\paragraph{Breadth and statistical rigor.} At scale, the recognition core remains strong across all $12$ domains and five modalities --- a total of $1{,}080$ tasks ($N\in\{10,20,40\}$, $30$ draws, $95\%$ CIs)---substantially outperforming random chance in every case (Figure~\ref{fig:crossdomain}). Importantly, recall \emph{tracks the dedicated encoder}: it is near-perfect where the encoder is strong (e.g., speaker and face identity), and lower where the encoder is weaker (e.g., style and tone). Notably, the VLM's own native vision tokens perform substantially worse than a purpose-built encoder on the same AgeDB faces. This provides clear evidence that the encoder --- not the memory mechanism --- determines the performance ceiling in each domain (\S\ref{sec:when}).

\paragraph{Training-free on any frozen model.} Because the read operation is a similarity-weighted blend of marker tokens, recall does not depend on the specific host language model and requires no training to transfer. We verify this across ten model families (Qwen2.5 1.5B/7B~\citep{qwen25}, Qwen3 4B/8B/14B~\citep{qwen3}, Phi-3.5~\citep{phi3}, SmolLM2~\citep{smollm2}, DeepSeek-Llama-8B~\citep{deepseekr1}, Mistral-7B~\citep{mistral7b}, and a hybrid-Mamba Granite~\citep{granite4}). In all cases, the in-model read reproduces the encoder's recall (Appendix Figure~\ref{fig:universality}). A five-architecture by five-modality grid shows the match holds to within one percentage point in every cell (Appendix Table~\ref{tab:univgrid}). On models with tied embeddings the agreement is within $0.003$; untied-embedding models match equally well with a modestly larger read-strength constant. Even the hybrid-Mamba architecture performs within one point. This universality is not the main contribution but a valuable supporting property: the storage mechanism can be hosted on \emph{any} model the agent already uses, at essentially zero additional cost.

\begin{figure}[t]
  \centering
  \includegraphics[width=0.88\linewidth]{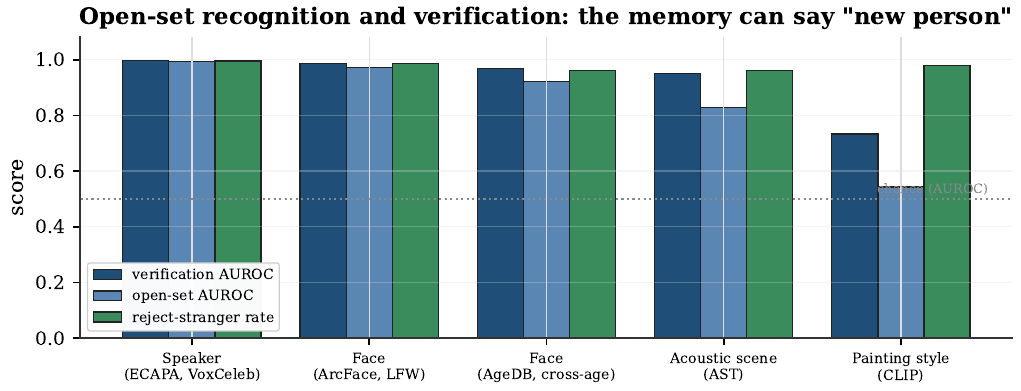}
  \caption{Beyond closed-set recall, the memory supports verification and open-set identification ($N{=}20$ enrolled, $30$ draws). For each modality we report verification AUROC (same vs different identity), open-set AUROC (enrolled vs never-enrolled probes), and the reject-stranger rate at the balanced threshold. Performance is high where the encoder is strong (voice, face) and lower where it is weak (painting style), mirroring the same trend observed in recall.}
  \label{fig:openset}
\end{figure}

\paragraph{Rejecting strangers and verifying identities.} A practical user memory must do more than closed-set recall, where the query is guaranteed to belong to a registered identity. It must also reject strangers it has never seen and accurately verify whether two perceptions belong to the same person. Both capabilities emerge naturally from the same read operation (cosine similarity over encoder keys). We evaluate these using standard operating-point metrics over $30$ draws (Figure~\ref{fig:openset}). On verification, the memory achieves strong separation between same- and different-identity pairs, with AUROC of $0.99$ on voice and $0.97$--$0.99$ on faces (equal error rates of $1.3\%$ and $5$--$8\%$ respectively). On open-set identification, it rejects never-enrolled strangers with $0.96$--$0.99$ accuracy while correctly identifying enrolled individuals. As in previous experiments, these results closely track the underlying encoder: performance is high where the encoder is strong (voice and face) and lower where it is weak (e.g., painting style, open-set AUROC $0.54$). The key takeaway is not new state-of-the-art numbers, but that our memory naturally supports the open-set recognition and verification modes required by a deployed companion agent, inheriting the encoder's separability in both settings.

\subsection{Composition with in-context text memory}
\label{sub:compose}

\begin{figure}[t]
  \centering
  \includegraphics[width=0.99\linewidth]{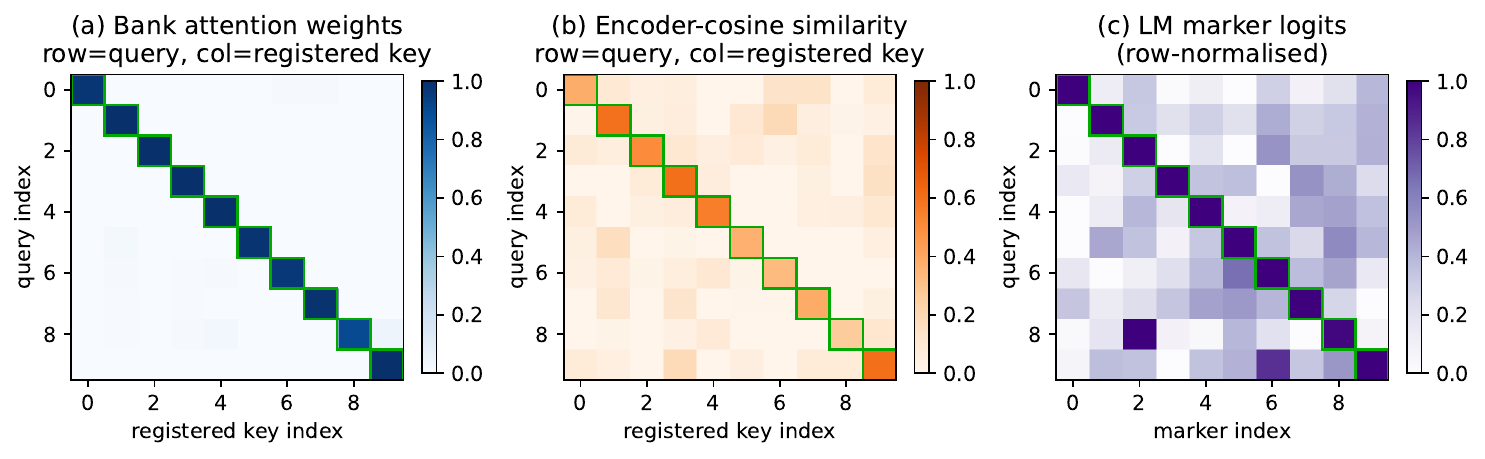}
  \caption{One recall, three views (held-out faces, ten identities). \textbf{(a)} With a sharp temperature, attention over the memory bank forms a near-pure diagonal (average $0.98$), strongly committing to the top encoder match. \textbf{(b)} The underlying encoder cosine similarities show a softer diagonal (avg.\ $0.46$) with noticeable off-diagonal mass; the memory faithfully inherits these similarities and their errors. \textbf{(c)} The resulting marker logits keep the correct identity at the top. Overall, the attention faithfully reads the encoder's nearest neighbour as a token; it does not sharpen the similarity.}
  \label{fig:mechanism}
\end{figure}

\begin{figure}[t]
  \centering
  \includegraphics[width=0.6\linewidth]{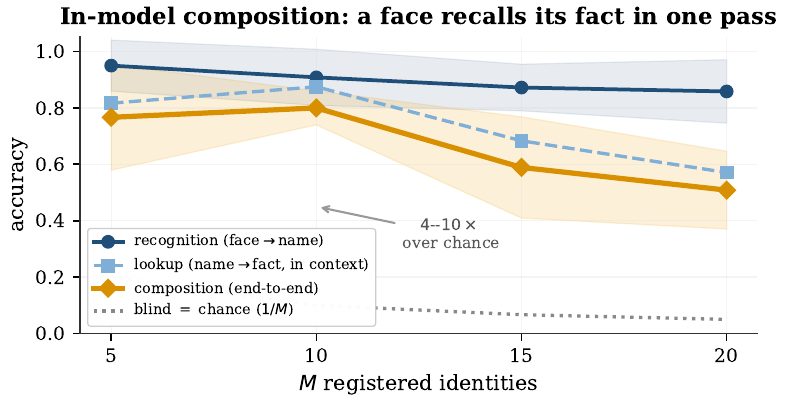}
  \caption{In-model composition: we register $M$ faces with dedicated name markers, bind one fact per name in context, then query the associated fact using a held-out cross-condition photo, all in one forward pass. End-to-end accuracy (gold) closely tracks the product of recognition accuracy (navy) times the model's in-context lookup accuracy (blue) and beats the face-withheld random chance baseline by $4$--$10\times$. Results averaged over 12 draws; bands show $\pm$ one standard deviation.}
  \label{fig:composition}
\end{figure}

The key advantage of recalling a perception as a native token (rather than as an external label) is that the token can participate directly in the model's reasoning. We demonstrate this by registering $M$ faces with dedicated name markers, then binding one fact per name in context (e.g., ``Anna is a teacher''). We then show a held-out cross-condition photo and ask the model to retrieve the associated fact. The entire chain runs in a single forward pass: the face recalls its name through \attmem{}, and the name retrieves the corresponding fact from context. End-to-end accuracy improves $4$--$10\times$ over the face-withheld chance baseline and closely tracks the product of recognition accuracy and the model's in-context lookup reliability (Figure~\ref{fig:composition}). As $M$ grows, the bottleneck shifts from the memory to in-context lookup. Figure~\ref{fig:mechanism} visualizes a single read: a sharp temperature concentrates the encoder's cosine similarities into a near-one-hot diagonal, producing decisive marker logits that match what pure encoder retrieval would select.

\subsection{Comparison with retrieval and in-context baselines}
\label{sub:cost}

The natural baseline is a retrieve-and-reprompt pipeline: cosine-retrieve the nearest registered face, recover its name, and re-prompt the model with the result. Because both approaches use the same encoder for recognition and the same frozen model for reasoning, they achieve \emph{identical accuracy} on both the composition task and a multi-perception question such as ``how many distinct people are here?'' (Table~\ref{tab:pipeline}). On pure recognition cost, the pipeline is actually cheaper, as it can count distinct identities using only cosine similarity with no model forward pass. We do not claim that the parametric memory is faster or more accurate at recognition itself. Its advantages are architectural. 

The pipeline is a two-pass system: an orchestrator must decide \emph{before} generation that recognition is needed, query an external index, and re-prompt the model. In contrast, the in-model read requires no such upfront decision. The memory bank is consulted automatically at every perceptual mention during the agent's normal generation. This enables seamless composition with ongoing reasoning (\S\ref{sub:compose}), supports interleaving multiple perceptions in a single stream, and ships as a single frozen-model artifact with no external index to manage, synchronize, or secure. When a deployment can reliably afford the extra pass, the pipeline is a reasonable choice. The parametric in-model memory is preferable for agents that cannot know in advance when a perception will require identification.

The comparison becomes even sharper against the baseline a VLM can actually achieve in practice: keeping the perception directly in context as image tokens. A single face crop consumes 64 vision tokens in Qwen2.5-VL at 224px resolution, and 256--576 tokens at higher resolutions (a full image approaches a thousand). In contrast, \attmem{} represents the same perception using just one marker token---a 64--576$\times$ reduction in context tokens per perception. Because the memory bank lives outside the context window, it scales without bound, whereas keeping images in context exhausts even a 128k window after roughly two thousand crops. On accuracy, the in-context image baseline is also substantially weaker: the VLM relies on its general-purpose vision tokens ($0.50$ end-to-end accuracy, Appendix Table~\ref{tab:vlm}), far below a dedicated face encoder ($0.95$). Thus, the parametric memory matches the performance of the strong external-index pipeline while significantly outperforming the in-context image baseline on both accuracy and cost. Combined with the capacity-law router, this is the core contribution of our approach.

\begin{table}[h]
\small\centering
\caption{In-model memory vs.\ a retrieve-and-reprompt pipeline (Qwen2.5-3B, 10 draws). Accuracy is identical in all cases, since both approaches use the same encoder cosine similarity for recognition. The pipeline is cheaper for pure recognition tasks (e.g., counting distinct identities requires no model forward pass). The in-model read, by contrast, folds recognition directly into the model's own forward pass, eliminating the need for an external index.}
\label{tab:pipeline}
\begin{tabular}{@{}lrrrr@{}}
\toprule
\textbf{Task} & \textbf{In-model acc} & \textbf{Pipeline acc} & \textbf{In-model lat.} & \textbf{Pipeline lat.} \\
\midrule
Compose (face$\to$fact)      & 0.81 & 0.81 & 122\,ms & 98\,ms \\
Count distinct, $K{=}2$      & 1.00 & 1.00 & 238\,ms & 0.09\,ms \\
Count distinct, $K{=}3$      & 1.00 & 1.00 & 361\,ms & 0.09\,ms \\
Count distinct, $K{=}4$      & 0.70 & 0.70 & 472\,ms & 0.09\,ms \\
\bottomrule
\end{tabular}
\end{table}

\subsection{Preservation of text-only behaviour}
\label{sub:nonreg}

The perceptual memory is designed to sit alongside---rather than interfere with---the model's ordinary text memory. The attention hook adds a residual only at perceptual positions, so the model's text-only behavior remains unchanged even when the memory bank is installed and populated. Concretely, with a populated bank the top-1 next-token predictions match the untouched model on every probe. The populated bank produces exactly the same logits as an empty one, and with the hook disabled the logits are bitwise identical (maximum $|\Delta\text{logit}| = 0$). The small residual deviation ($\approx 0.375$) observed on the wrapped forward pass stems from the host model's \texttt{inputs\_embeds} numerics and appears regardless of whether the bank is empty or full. Registering faces and voices together shows the two banks remain fully independent with no training (cross-modal leakage of $0.017$ for faces and $0.067$ for voices). An agent can therefore store factual knowledge such as ``my favourite restaurant is in Paris'' in its text memory and perceptual identity such as ``how Bibi looks across lighting and age'' in its perceptual bank without compromising either. For contrast, we also evaluated a discrete codebook design (similar to previous \patha{}-style approaches) that maps each perception to one of $K$ learned codes. As expected, it never exceeded $\sim$0.07 recall beyond a few hundred identities---the same fundamental limitation as captioning---because it discards fine-grained signal that the continuous encoder preserves. These results are included in Appendix~\ref{app:patha}.

\subsection{An end-to-end multi-session agent}
\label{sub:agent}

\begin{figure}[t]
  \centering
  \includegraphics[width=0.72\linewidth]{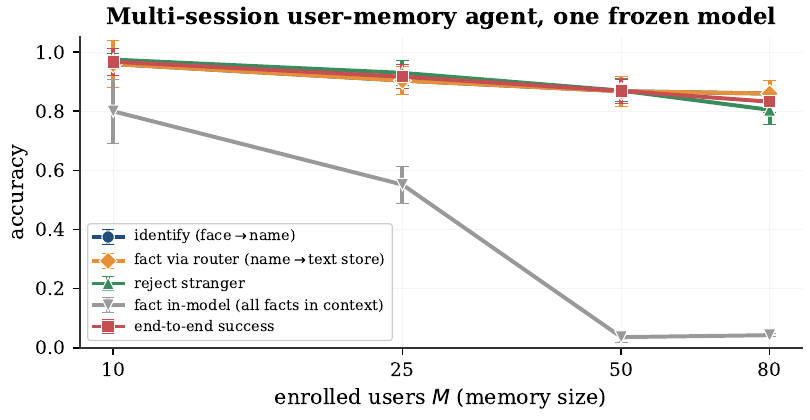}
  \caption{An end-to-end multi-session user-memory agent (Qwen2.5-3B, 5 seeds) as the enrolled population grows from 10 to 80 users. The system identifies returning users (face$\to$name), recalls associated facts via the router (name$\to$text store), and correctly rejects strangers. All capabilities scale gracefully, tracking the underlying encoder. In contrast, attempting to recall facts \emph{in-model} by binding all of them in context collapses beyond a few dozen users --- limited by the host model's context window, not the memory itself. This demonstrates why the router keeps exact facts in a text store. Overall end-to-end task success remains strong at $0.83$ with eighty enrolled users, all within a single frozen model.}
  \label{fig:agent}
\end{figure}

The components come together in a complete working agent. We construct a system consisting of a frozen language model equipped with the \attmem{} perceptual memory bank and a separate text store for facts. Users are enrolled across sessions, each with a face (registered from one photo) and an associated fact. When a returning user appears in a new cross-condition photo, the agent must identify them, answer a question about them, while also correctly rejecting anyone it has never met. We evaluate performance as the enrolled population grows from 10 to 80 users, averaged over five seeds (Figure~\ref{fig:agent}). The perceptual capabilities scale gracefully: identification accuracy remains at $0.86$ and stranger rejection at $0.81$ even with 80 enrolled users, closely tracking the encoder (as observed throughout). Note that the higher rejection rates reported in \S\ref{sub:core} ($0.96$--$0.99$) were measured at only 20 enrolled users; as the number of identities increases, the operating point naturally tightens in line with the encoder's separability.

Fact recall highlights the value of the router. When perceptual identity is routed to the parametric bank (which recalls the name) and the name then keys an exact lookup in the text store, accuracy tracks identification performance ($0.86$ at 80 users). In contrast, attempting to bind all facts in the prompt and resolve everything in a single \emph{in-model} pass collapses beyond a few dozen users ($0.55$ at 25 users, $0.04$ at 50 users), because even a 3B model cannot reliably hold that many name--fact bindings in context. This provides a concrete demonstration of the router thesis: perceptual identity belongs in the parametric bank, where it scales effectively, while exact facts belong in a text store, where lookup remains precise. The resulting production path---perceptual bank for \emph{who}, text store for \emph{what}---achieves strong end-to-end task success ($0.83$ at 80 users) entirely within a single frozen model, with no external index required.

\section{The encoder determines recall}
\label{sec:when}

Because the in-model read faithfully reproduces the encoder's recall, the overall quality of the memory system is determined by the quality of the encoder. On the exact same identities, recall tracks encoder strength closely: on AgeDB, ArcFace reaches $0.95$ recall at ten identities and $0.75$ at a hundred, whereas the VLM's own vision tokens manage only $0.64$ and $0.34$ respectively on those same faces. A learned similarity metric trained on the same features performs no better (Appendix~\ref{app:tables}). This is the fundamental reason we use a dedicated encoder for the identification step rather than relying on the model's native tokens. The performance gap persists even when running the full system end-to-end inside a real VLM (\S\ref{sub:cost}, Appendix Table~\ref{tab:vlm}). In contrast, the storage mechanism itself is model-agnostic: the read operation contains no encoder-specific machinery and ports to any frozen language model without training (\S\ref{sub:core}). If better perceptual recall is needed for a task, the only effective solution is a stronger encoder --- not a different language model or a more sophisticated read mechanism.

\section{Related work}
\label{sec:related}

\textbf{Text-first agent memory.} The conversational-memory line --- Mem0~\citep{mem0}, MemGPT and its Letta successor~\citep{memgpt,letta}, MemoryLLM~\citep{memllm}, M3-Agent~\citep{m3agent}, benchmarked by LongMemEval~\citep{longmemeval} --- stores transcripts and captions in a text index and retrieves by similarity. These systems define the captionable half of user memory that ours composes with. None of them keep a perception in its native modality, which is the missing half this paper adds (\S\ref{sub:textcant} measures what that omission costs).

\textbf{Personalized vision-language models.} MyVLM~\citep{myvlm}, Yo'LLaVA~\citep{yollava}, MC-LLaVA~\citep{mcllava}, RAP~\citep{rap}, Online-PVLM~\citep{onlinepvlm}, and TAME~\citep{tame} personalize a VLM to user-specific concepts, typically by learning an artifact per concept --- a classifier head, a soft token, a projection. Our setting differs in both cost profile and mechanism: registration is a training-free $\mathcal{O}(1)$ row append. As shown in Appendix Table~\ref{tab:perconcept}, these learned per-concept artifacts do not outperform the encoder's own cosine similarity on cross-condition identity tasks, while incurring per-identity gradient updates.

\textbf{Retrieval inside the model.} \attmem{} adopts the general shape of attention over an external bank from $k$NN-LM~\citep{khandelwal2020nearest}, Memorizing Transformers~\citep{wu2022memorizing}, and RETRO~\citep{retro}, which retrieve from token or chunk stores to improve language modeling. We repurpose this mechanism for a different goal: the keys are perceptual encoder embeddings rather than text hidden states, the bank stores a user's persistent perceptions rather than a corpus, and the read lands at the output head of a frozen model with no training.

\textbf{Perception-to-token bridges and speaker recognition.} Models such as Flamingo's resampler~\citep{flamingo} and BLIP-2's Q-Former~\citep{blip2} learn to project perception into a language model's token space. In contrast, our approach injects a single marker token whose value is the model's \emph{own} embedding, which is what enables training-free operation. The speaker recognition and diarization literature~\citep{voxceleb,ecapa,diarizationreview} treats ``who spoke when'' as a standalone pipeline that returns labels. We instead use the same encoders to produce grounded identity keys that the model can directly condition on, rather than treating them as external verdicts.

\section{Discussion}
\label{sec:discussion}

\textbf{The claim is structural, and the numbers illustrate it.} Our central argument is not that ``our method scores higher.'' Rather, it is that text is the wrong container for a large and important portion of what an agent should remember about a person, that recognizing a perception naturally decomposes into two distinct problems---grounding the referent in context and identifying it---and that a frozen model fronted by a VLM and backed by a dedicated encoder bank can solve both halves, whereas text and a bare encoder each solve only one (or neither).

The empirical results substantiate each piece of this argument: the performance penalty of text is real and strongly modality-dependent (Figure~\ref{fig:textablation}); grounding is both necessary and highly effective at recovering oracle-level recognition in cluttered scenes (Figure~\ref{fig:agentic}); and the recognition core is the dedicated encoder itself, given an in-model home that is training-free and model-agnostic (Appendix Figure~\ref{fig:universality}). The core contribution is therefore the grounded, in-model perceptual memory container---not a particular recall number. The numbers we report simply reflect the quality of the underlying encoder.

\textbf{This is additive infrastructure, not a replacement.} The perceptual memory is designed to bolt onto the text-based memory systems that agents already use~\citep{mem0,memllm,memgpt,letta}. The strong non-regression results support this integration: text-only behavior is preserved exactly, and the perceptual bank activates only at perceptual mentions. As long-term memory benchmarks such as LongMemEval~\citep{longmemeval} begin to incorporate perceptual content, the structural gap highlighted in this work will become directly measurable.

\textbf{The captionable half is going parametric too.} This paper is one half of a broader shift: a personalized agent should keep its memory \emph{inside} the model rather than in an external index. A parallel line of work is making the same architectural choice for the \emph{captionable} (textual/factual) half of user memory. \emph{User as Engram}~\citep{userengram} stores per-user facts in a hash-keyed parametric memory and shows that the intuitive alternative---a per-user LoRA adapter~\citep{lora}---is \emph{reasoning-negative}: it successfully memorizes facts but degrades the model's general reasoning, because LoRA edits are global while row insertions are local. \emph{User as Code}~\citep{usercode} encodes user information as executable typed state, and \emph{Programmable KV Cache}~\citep{editablekv} maintains editable, composable notes directly in the attention cache.

Our perceptual memory shares that thesis with a different mechanism: continuous cross-attention over encoder banks rather than hash-keyed rows. This continuous channel, which carries a perception as a vector projected into the model's own embedding space (rather than as text), aligns with the approach taken by \emph{The Latent Bridge}~\citep{latentbridge} for inter-model communication, where a learned latent link outperforms writing and re-reading text. Together, the two halves compose into one memory system that holds both what the user said and what they are like.

\begin{figure}[t]
  \centering
  \includegraphics[width=0.95\linewidth]{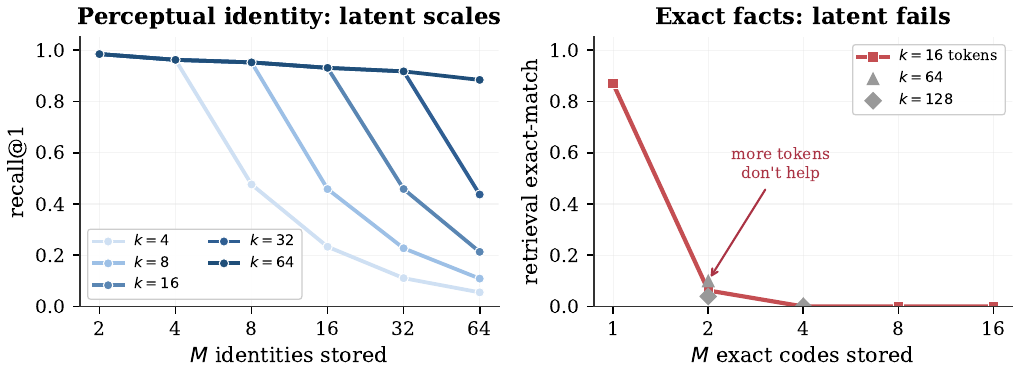}
  \caption{Two capacity laws for memory compressed into $k$ soft tokens read by a frozen language model, the mechanism behind the router. \textbf{Left:} Perceptual identity is capacity-limited. When compressing $M$ faces into $k$ prototype slots and querying with a cross-condition sample, recall scales as $\approx\min(1, k/M)\times C(M)$ where $C(M)$ is the encoder ceiling ($\approx 1$ for faces). Voices follow the same pattern. One slot per identity recovers full encoder performance, and more slots allow more identities to be stored. \textbf{Right:} Exact factual content is binding-limited. A latent can reliably hold and retrieve a single exact code ($0.87$), but attempting to store and retrieve multiple codes collapses immediately ($0.06$ at two codes), and adding more latent tokens does not help. Recognition scales with the latent budget while exact recall does not, which is why it belongs in a text store.}
  \label{fig:capacity}
\end{figure}

\textbf{The router: what a latent memory can and cannot hold.} The split between the parametric perceptual bank and the text store is not an arbitrary convention. It follows from two capacity laws that we measured by compressing content into $k$ soft tokens read by a frozen language model (Figure~\ref{fig:capacity}). Perceptual identity is capacity-limited and degrades gracefully. Compressing $M$ registered identities into $k$ prototype slots and querying with a cross-condition sample, the system can distinguish roughly a $k/M$ fraction of them. This holds consistently across faces, voices, acoustic scenes, painting styles, and tones of voice (Appendix Table~\ref{tab:capacitygrid}). With one slot per identity, performance matches the encoder; with fewer slots, identities are lost proportionally. The general law is recall $\approx \min(1, k/M)\cdot C(M)$, where $C(M)$ is the encoder's own recall at $M$ registrations. The slot-compression factor $\min(1,k/M)$ is universal, while the ceiling $C(M)$ is determined by the encoder (near $1$ for strong encoders such as face and voice, lower for weaker ones such as style and tone).

This behavior is not an artifact of $k$-means clustering: a compressor whose $k$ slots are learned via gradient descent follows the same curve, as $k$ hard slots can separate at most $k$ out of $M$ identities (a pigeonhole bound). Exact factual content behaves in the opposite way. A latent representation can hold and retrieve a single exact code reliably ($0.87$ for a six-character code), but attempting to store multiple codes and retrieve one by name collapses immediately ($0.06$ at two codes, near zero beyond). Adding more latent tokens does not help. Recognition is fundamentally a capacity problem (one identity per slot), while exact recall is a binding problem (matching a key to its precise value), which a latent memory cannot solve at any practical budget. This is why the captionable half belongs in a text store, which performs exact lookup, and the perceptual half belongs in a parametric bank, which scales with the number of slots. This boundary matches exactly what the text ablation showed empirically (Figure~\ref{fig:textablation}), now derived from capacity considerations.

\textbf{Privacy and responsible use.} A perceptual user memory stores biometric templates (faceprints and voiceprints) and must be held to the high standards required for such data. Enrollment should be consented and explicit (the ``remember her'' act of Figure~\ref{fig:worked} is an affirmative request, not passive collection). The bank should be stored on-device or encrypted at rest, and the system must not be repurposed to identify individuals who have not consented to being remembered. Two properties of our design support responsible deployment. First, because each identity occupies a single row, deletion is as simple as registration: removing the row completely erases the identity with no fine-tuned weights to unlearn. This provides a clean implementation of the right to be forgotten, which per-concept tuning approaches cannot easily offer. Second, because recall performance is determined by the underlying encoder, the memory inherits the encoder's known demographic biases (particularly in face recognition). Deployments should therefore carefully evaluate the chosen encoder on their target population and favor verification-style operating points with strong stranger rejection enabled (Figure~\ref{fig:openset}), ensuring that unenrolled individuals are refused rather than misassigned to an enrolled identity.

\textbf{Limitations.}
\begin{enumerate}
  \item \textbf{The memory cannot exceed the encoder.} Its recall matches that of the underlying encoder --- paired to within $\pm0.001$ from $N{=}5$ to $N{=}1000$ (Table~\ref{tab:headline}, Figure~\ref{fig:scaling}). Consequently, performance at very large scale declines exactly as the encoder's cosine similarity does ($0.78$ at a thousand faces), and no in-model mechanism can recover more.
  \item \textbf{Scale of cross-condition evaluation.} Our largest cross-condition pool holds $2{,}180$ real identities, from which we measure recall to $N{=}1000$ registered users. Constant-time behaviour is confirmed up to ten thousand rows, but recall at that scale remains to be tested with larger datasets. 
  \item \textbf{Grounding robustness.} The grounding step inherits the referring-expression resolution capabilities of the VLM. While it is near-lossless on well-separated composites (grounding accuracy $1.00$ up to $K{=}5$), performance degrades under more realistic stress conditions. For example, jittered and rescaled faces pasted onto photographic backgrounds reduce grounding accuracy to $0.63$--$0.81$ and end-to-end recall well below the oracle (Appendix Table~\ref{tab:realistic}). Audio grounding of the fourth speaker drops to $0.69$. Denser, occluded, or noisier scenes will stress it further, making detector and diarization robustness---rather than the encoder---the practical bottleneck.
  \item \textbf{Dependence on external encoders.} Inside a VLM, native vision tokens are weak identity keys ($0.50$ vs.\ ArcFace's $0.95$ at $N{=}10$, Table~\ref{tab:vlm}). A dedicated, external modality-specific encoder is therefore required.
  \item \textbf{Per-model hyperparameter.} The one per-model constant (read-strength gain) is set by inspection. Untied-embedding models need a larger value, which we have not automated. 
  \item \textbf{Baselines.} The closest prior systems (MyVLM~\citep{myvlm}, Online-PVLM~\citep{onlinepvlm}) have released neither code nor checkpoints. Comparisons are therefore made against charitable reimplementations of their core mechanisms (Appendix~\ref{app:tables}, Table~\ref{tab:perconcept}).
\end{enumerate}

\section{Conclusion}
\label{sec:conclusion}

A photograph and a written description of a face are not two encodings of the same thing. The description has already discarded much of what enables recognition. Today's agent memory systems keep only the description. We have argued that they must also keep the photograph itself---and that doing so effectively requires grounding. A vision-language model resolves the referent the user means in context. A dedicated encoder turns that referent into an identity that survives variations text cannot capture. A frozen language model then stores this identity as a content-addressable row, read directly inside its own forward pass. Each component addresses what the others cannot: the VLM cannot reliably identify, the encoder cannot ground, and an external index cannot return the answer as a native token inside the model. Together, they achieve oracle-level recognition that none of them can deliver alone. The recognition performance itself belongs to the encoder; our contribution is giving it an effective home---training-free, constant-time, and portable to any frozen model. The two memory systems are complementary: text for what the user \emph{said}, and parametric multimodal memory for what the user \emph{is like}. An agent equipped with both strictly dominates one that has only text. This is where a face stops being reduced to a sentence, and starts becoming a memory.

\section*{Acknowledgements}

We thank Xiangyu Zhang from StepFun AI for the initial ideas that inspired this paper; he introduced the problem of multimodal user memory to the authors and pointed out the limitations of purely textual memory.
This paper was produced using Pine Copilot's voice-directed \emph{whisper coding} workflow~\citep{pineai2026whispercoding}, in which the authors specify, discuss, and review the work by voice while a coding agent (Claude Code with Claude Opus 4.8) carries out the planning, coding, experiments, and paper writing.
We thank BSQL Networking for hosting the NVIDIA RTX PRO 6000 GPU.

\appendix

\section{Detailed results}
\label{app:tables}

This appendix collects the per-cell numbers behind Sections~\ref{sec:results}--\ref{sec:when}. Recall is top-1 with the argmax restricted to registered markers. Tables were measured on different face pools, splits, and draw protocols (single-draw sweeps versus the 20-draw paired protocol below), so raw-cosine columns differ slightly across tables (e.g.\ $0.767$ vs.\ $0.776$ at $N{=}1000$); comparisons are valid within a table, not across tables. Tables marked \emph{single eval draw} (Tables~\ref{tab:learnedmetric} and~\ref{tab:perconcept}) are directional comparisons on one shared draw; the paired multi-draw protocol below is the statistical reference.

\paragraph{The 12 PerceptMem domains.} Table~\ref{tab:domains} enumerates the benchmark behind Figure~\ref{fig:crossdomain}: each domain is a (dataset, encoder) pair, evaluated at $N\in\{10,20,40\}$ with $30$ draws each ($90$ tasks per domain, $1{,}080$ total). The five conceptual sub-modalities of Table~\ref{tab:datasets} each appear under at least one dataset--encoder pairing, several under two, plus the VLM's native vision tokens as the deliberately weak twelfth key space.

\begin{table}[h]
\small\centering
\caption{The $12$ PerceptMem domains: dataset, identity key encoder, cross-condition identity pool, and parametric recall@1 at $N{=}20$ (mean $\pm$95\% CI over $30$ draws; chance $0.05$). Tone identities are (speaker, emotional-state) pairs. The Qwen2.5-VL native-token row is included as a key-space control, not a recommended configuration (\S\ref{sec:when}).}
\label{tab:domains}
\begin{tabular}{@{}llrr@{}}
\toprule
\textbf{Domain} & \textbf{Key encoder} & \textbf{Pool} & \textbf{Recall@1, $N{=}20$} \\
\midrule
Face / AgeDB (cross-age)        & ArcFace                & 500  & $0.869 \pm 0.021$ \\
Face / LFW                      & ArcFace                & 1680 & $0.932 \pm 0.018$ \\
Face / LFW                      & AntelopeV2             & 901  & $0.953 \pm 0.012$ \\
Face / LFW$+$AgeDB combined     & ArcFace                & 2180 & $0.921 \pm 0.019$ \\
Speaker / LibriSpeech           & ECAPA-TDNN             & 58   & $0.992 \pm 0.004$ \\
Speaker / VoxCeleb (in-wild)    & ECAPA-TDNN             & 40   & $0.991 \pm 0.005$ \\
Acoustic scene / ESC-50         & AST                    & 50   & $0.792 \pm 0.023$ \\
Painting style / WikiArt        & CLIP (mid-layer)       & 128  & $0.319 \pm 0.024$ \\
Painting style / WikiArt        & DINOv2~\citep{dinov2}  & 50   & $0.236 \pm 0.023$ \\
Vocal tone / RAVDESS            & wav2vec2 (emotion)     & 168  & $0.496 \pm 0.021$ \\
Face / AgeDB (control)          & Qwen2.5-VL native tokens & 567 & $0.542 \pm 0.029$ \\
Painting style / WikiArt        & contrastive head       & 26   & $0.245 \pm 0.018$ \\
\bottomrule
\end{tabular}
\end{table}

\paragraph{Paired evaluation protocol.} A comparison between the memory and retrieval requires two controls: both methods score on \emph{identical} registrations and queries, and each target identity occupies a randomly chosen bank slot. The second control matters because a fixed target slot lets the constant marker token in that slot accrue a baseline output-logit, which inflates recall independently of the query. We score both methods on the same twenty draws with the target slot randomised, and report mean$\pm$std. Table~\ref{tab:headline} gives the result, with random-bank face rows spanning $N{=}10$ to $N{=}1000$. The memory ties retrieval at every size and in every regime, since with a sharp read it computes the same argmax over encoder cosine --- even on banks of the nineteen most-confusable look-alikes, where the cosine gaps are smallest ($0.853$ vs.\ $0.853$, $\Delta{=}0.0$pp). Recall is not monotone in $N$ ($0.749$ at $N{=}300$ vs.\ $0.776$ at $N{=}1000$): a draw at $N{=}1000$ spans nearly the whole $1090$-identity eval pool, whereas smaller $N$ are random subsets whose average difficulty differs from the pool's; the effect is identical in both columns, so it is a property of the encoder and the pool, not of the memory. The sharpness is load-bearing: an earlier trained variant whose learned temperature settled at inv-temp $\approx 20$ trailed retrieval by eight points on the look-alike cell and by nine to eighteen points at $N\ge300$, because a soft blend of marker values carries no per-identity signal beyond the cosine it dilutes. With the constants of Section~\ref{sec:method} the blend is a faithful argmax in every cell we measure.

\begin{table}[h]
\small\centering
\caption{Paired evaluation: the memory and retrieval scored on identical registrations and queries across 20 draws, with the target slot randomised. The memory ties retrieval everywhere --- random banks through $N{=}1000$, style banks, and adversarial banks of the nineteen most-confusable look-alikes. Recall is the encoder's throughout.}
\label{tab:headline}
\begin{tabular}{@{}llrrr@{}}
\toprule
\textbf{Regime} & \textbf{Cell} & \textbf{Retrieval} & \textbf{\attmem{} (mean$\pm$std)} & \textbf{$\Delta$} \\
\midrule
random      & Face, $N{=}10$        & $0.948$ & $0.948 \pm 0.033$ & $0.0$pp \\
random      & Face, $N{=}100$       & $0.792$ & $0.792 \pm 0.021$ & $0.0$pp \\
random      & Face, $N{=}300$       & $0.749$ & $0.749 \pm 0.018$ & $0.0$pp \\
random      & Face, $N{=}1000$      & $0.776$ & $0.777 \pm 0.008$ & $0.0$pp \\
random      & Style, $N{=}5$        & $0.473$ & $0.473 \pm 0.104$ & $0.0$pp \\
random      & Style, $N{=}10$       & $0.428$ & $0.428 \pm 0.076$ & $0.0$pp \\
adversarial & Face, $K{=}19$        & $0.853$ & $0.853 \pm 0.007$ & $0.0$pp \\
\bottomrule
\end{tabular}
\end{table}

\paragraph{Universality grid.} Table~\ref{tab:univgrid} runs the training-free read across five model architectures and all five modalities, scoring each cell against the encoder with the same paired protocol (20 draws). Every one of the twenty-five cells reproduces the encoder's recall to within one point (worst case $0.010$, most cells $\le 0.003$). Tied-embedding models use the default read-strength gain; untied and hybrid models use a single larger value ($256$), set once per model and held fixed across modalities. The read therefore inherits the encoder's recall regardless of the host model's family, embedding scheme, scale, or even whether it is a transformer.

\begin{table}[h]
\small\centering
\caption{Universality grid: maximum $|\Delta|$ between the training-free memory and the encoder over $N\in\{5,10,20\}$ (paired, 20 draws), for five architectures $\times$ five modalities. Every cell is within one point of the encoder. ``gain'' is the single per-model read-strength constant. Encoder recall@1 at $N{=}10$: Face $0.95$, Speaker $0.99$, Acoustic $0.86$, Tone $0.52$, Style $0.43$.}
\label{tab:univgrid}
\begin{tabular}{@{}llrrrrrr@{}}
\toprule
\textbf{Model} & \textbf{Emb.} & \textbf{gain} & \textbf{Face} & \textbf{Speaker} & \textbf{Acoustic} & \textbf{Tone} & \textbf{Style} \\
\midrule
Qwen2.5-7B & tied & 64 & 0.001 & 0.000 & 0.002 & 0.003 & 0.001 \\
Phi-3.5-mini & tied & 64 & 0.001 & 0.000 & 0.000 & 0.002 & 0.002 \\
DeepSeek-Llama-8B & untied & 256 & 0.000 & 0.000 & 0.002 & 0.002 & 0.000 \\
Mistral-7B & untied & 256 & 0.000 & 0.000 & 0.003 & 0.010 & 0.000 \\
Granite-4.0 (Mamba) & hybrid & 256 & 0.003 & 0.000 & 0.003 & 0.001 & 0.000 \\
\bottomrule
\end{tabular}
\end{table}

\begin{figure}[h]
  \centering
  \includegraphics[width=0.92\linewidth]{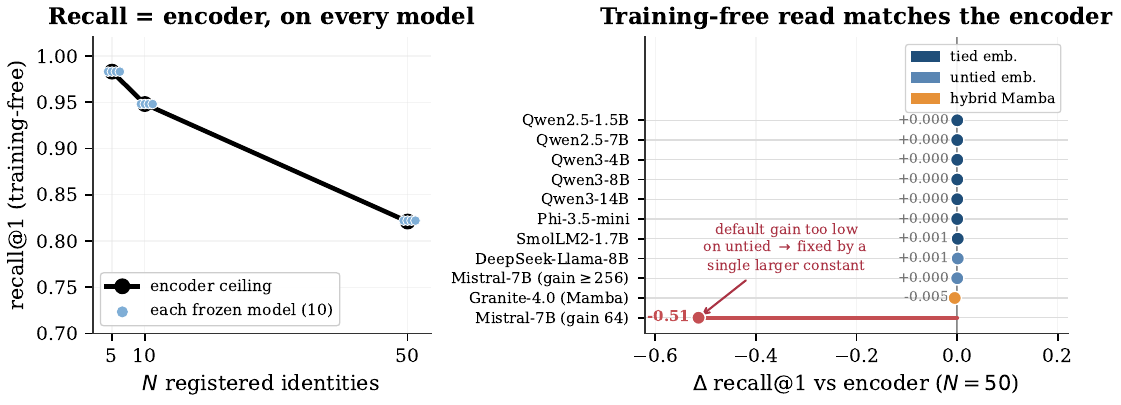}
  \caption{The training-free read reproduces the encoder on every frozen model, the supporting property behind the ``store'' step. \textbf{Left:} each model's training-free recall lands on the encoder ceiling (paired, 20 draws). \textbf{Right:} the gap from the encoder at $N{=}50$ is $\approx 0$ across ten families (1.5B--14B, tied and untied embeddings, hybrid-Mamba); only an untied model at the default gain dips (red), fixed by a single larger read-strength constant. No gradient steps anywhere.}
  \label{fig:universality}
\end{figure}

\paragraph{Read-strength robustness.} The one per-model constant, the residual gain, is set by inspection rather than search because exact recall holds across a wide range (Table~\ref{tab:gainrobust}, face $N{=}50$, paired). A tied-embedding model is exact at every gain from $16$ to $1024$; an untied one (Mistral) is below the encoder until a threshold near $128$--$256$ and then exact and stable through $1024$. Any value in the upper range works for either, so there is no tuning to get wrong.

\begin{table}[h]
\small\centering
\caption{Read-strength robustness: training-free recall@1 (face, $N{=}50$, encoder ceiling $0.821$) as the residual gain sweeps $16$--$1024$. Tied models are exact throughout; untied models are exact above a threshold and stay exact. The constant tolerates a wide range.}
\label{tab:gainrobust}
\begin{tabular}{@{}lrrrrrrr@{}}
\toprule
\textbf{Model} & \textbf{16} & \textbf{32} & \textbf{64} & \textbf{128} & \textbf{256} & \textbf{512} & \textbf{1024} \\
\midrule
Qwen2.5-3B (tied)   & 0.821 & 0.821 & 0.821 & 0.821 & 0.821 & 0.821 & 0.821 \\
Mistral-7B (untied) & 0.022 & 0.060 & 0.307 & 0.768 & 0.821 & 0.821 & 0.821 \\
\bottomrule
\end{tabular}
\end{table}

\paragraph{Capacity law across modalities.} Table~\ref{tab:capacitygrid} runs the perceptual capacity probe (compress $M$ registered identities into $k$ prototype slots, recognise a cross-condition query, 5 seeds) on all five modalities. The slot-compression factor $\min(1,k/M)$ fits every modality once normalised by the one-slot-per-identity ceiling $C(M)$: the normalised error is $\le 0.012$ on the strong-encoder modalities and $\le 0.13$ on the weaker style and tone encoders. The ceiling $C(M)$ itself is the encoder's recall and varies accordingly, from near $1$ on voice to $0.26$--$0.36$ at $M{=}32$ on style and tone. Recognition is capacity-limited and graceful in every modality; only the ceiling moves.

\begin{table}[h]
\small\centering
\caption{Perceptual capacity law per modality. $C(M)$ is the one-slot-per-identity ceiling (recall at $k{=}M$), i.e.\ the encoder's recall at $M$. ``fit'' is the mean $|\,\text{recall}/C(M) - \min(1,k/M)\,|$ over the $k{<}M$ cells: the slot-compression factor is universal; the ceiling is the encoder's.}
\label{tab:capacitygrid}
\begin{tabular}{@{}lrrrr@{}}
\toprule
\textbf{Modality} & $C(2)$ & $C(8)$ & $C(32)$ & \textbf{fit to $\min(1,k/M)$} \\
\midrule
Face (ArcFace)    & 0.98 & 0.95 & 0.92 & 0.002 \\
Voice (ECAPA)     & 1.00 & 1.00 & 0.99 & 0.001 \\
Acoustic (AST)    & 0.95 & 0.87 & 0.76 & 0.012 \\
Style (CLIP)      & 0.76 & 0.46 & 0.26 & 0.070 \\
Tone (wav2vec)    & 0.94 & 0.67 & 0.36 & 0.125 \\
\bottomrule
\end{tabular}
\end{table}

\paragraph{Learned-metric baseline.} A natural question is whether \emph{any} re-encoding of the same features beats raw cosine, even if our read does not. We fit a similarity on the same encoder features and the same identities (the 50/50 split, seed 42) and score it with the identical protocol. Two closed-form metrics suffice: regularised LDA and within-class whitening. Table~\ref{tab:learnedmetric} shows LDA edging raw cosine at small memory and falling below it past a few hundred identities, while whitening never moves more than a point from raw cosine in either direction. So there is no meaningful discrimination left to extract from these features: the encoder's cosine is already the best ruler, which is why the memory's inheriting it exactly is the ceiling, not a shortfall. A contrastive linear head we also trained underperformed both closed-form metrics and is omitted as a weak instantiation.

\begin{table}[h]
\small\centering
\caption{Learned-metric baseline, recall@1, same encoder, same split, all methods scored on the same eval draw. LDA edges raw cosine at small $N$ and falls below it at scale; whitening stays within a point of raw cosine throughout. No re-encoding of the same features meaningfully beats the cosine. \attmem{}'s recall equals raw cosine (paired through $N{=}1000$, Table~\ref{tab:headline}), so the raw-cosine column doubles as its recall.}
\label{tab:learnedmetric}
\begin{tabular}{@{}llrrr@{}}
\toprule
\textbf{Modality} & $N$ & \textbf{Raw cosine ($=$ \attmem{})} & \textbf{LDA} & \textbf{Whitening} \\
\midrule
Face (ArcFace)  & 5    & 0.933 & \textbf{1.000} & 0.933 \\
Face (ArcFace)  & 10   & 0.933 & \textbf{0.967} & 0.933 \\
Face (ArcFace)  & 300  & 0.734 & 0.667 & \textbf{0.744} \\
Face (ArcFace)  & 1000 & 0.767 & 0.724 & \textbf{0.776} \\
\midrule
Style (CLIP-mid) & 5   & 0.400 & 0.200 & \textbf{0.600} \\
Style (CLIP-mid) & 10  & \textbf{0.400} & 0.333 & 0.367 \\
\bottomrule
\end{tabular}
\end{table}

\paragraph{Comparison with per-concept methods.} The personalized-VLM line (MyVLM, Yo'LLaVA, Online-PVLM) registers a concept by \emph{learning} something per concept (a classifier head, a soft token, a projection). Their code and checkpoints are unreleased, so we reimplement each method's core mechanism charitably on the cross-condition face task (ArcFace keys, same split). Table~\ref{tab:perconcept} reports accuracy and registration cost. Two findings. First, a per-concept linear classifier (MyVLM's mechanism) is indistinguishable from raw cosine at every memory size (within single-draw noise), because with one registration embedding per identity the trained head collapses to the embedding itself. It yields no accuracy gain over retrieval, and it costs per-identity gradient descent that grows to $4.5$\,s at a thousand identities, against our $\mathcal{O}(1)$ tensor append. Second, a learned projection trained with supervised contrastive loss (Online-PVLM's mechanism) \emph{underperforms} raw cosine by $15$ to $28$ points on this task. We read this as a caution that a learned re-encoding is not without cost: tuned on identity-disjoint data, it can distort the very cross-condition geometry it was meant to sharpen. No method here, ours included, beats raw cosine on accuracy at scale; the parametric memory's case rests on cost and in-model behaviour, not recall.

\begin{table}[h]
\small\centering
\caption{Charitable reimplementations of per-concept methods on cross-condition face recall (ArcFace, recall@1, single eval draw). MyVLM's per-concept classifier ties raw cosine but its registration cost grows with the memory; Online-PVLM's learned projection underperforms raw cosine. \attmem{}'s recall equals raw cosine (paired through $N{=}1000$, Table~\ref{tab:headline}) at $\mathcal{O}(1)$ registration cost, so the raw-cosine column doubles as its recall.}
\label{tab:perconcept}
\begin{tabular}{@{}rrrrr@{}}
\toprule
$N$ & \textbf{Raw cosine ($=$ \attmem{})} & \textbf{MyVLM} & \textbf{Online-PVLM} & \textbf{MyVLM reg.\ cost} \\
\midrule
5    & 0.933 & 0.933 & 0.800 & 0.009\,s \\
10   & 1.000 & 0.967 & 0.750 & 0.019\,s \\
50   & 0.767 & 0.753 & 0.675 & 0.178\,s \\
100  & 0.780 & 0.770 & ---   & 0.528\,s \\
300  & 0.728 & 0.725 & ---   & 1.39\,s \\
1000 & 0.776 & 0.778 & ---   & 4.52\,s \\
\bottomrule
\end{tabular}
\end{table}

\begin{figure}[h]
  \centering
  \begin{subfigure}{0.46\linewidth}\centering
    \includegraphics[width=\linewidth]{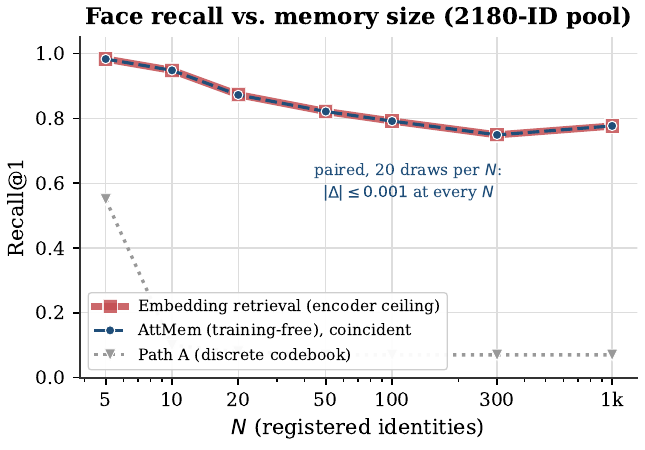}
    \caption{scaling on the face pool}
    \label{fig:scaling}
  \end{subfigure}\hfill
  \begin{subfigure}{0.46\linewidth}\centering
    \includegraphics[width=\linewidth]{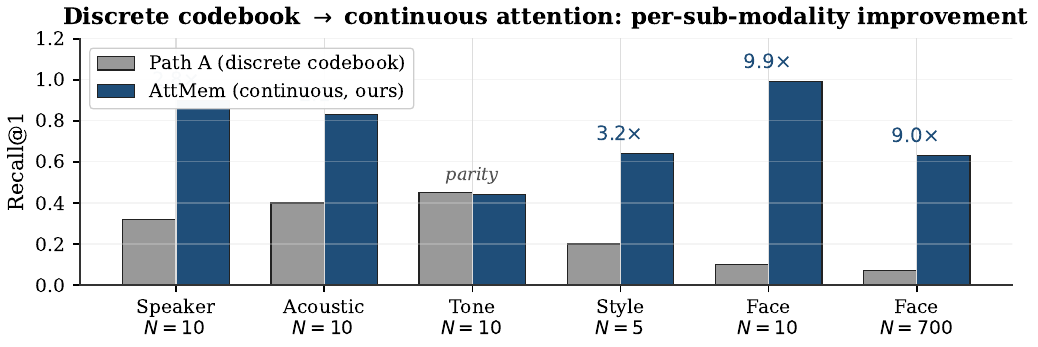}
    \caption{codebook vs.\ continuous attention}
    \label{fig:pivot}
  \end{subfigure}
  \caption{Left: under the paired protocol ($20$ draws per $N$) the training-free read is coincident with the encoder ceiling from $N{=}5$ to $N{=}1000$ ($|\Delta|\le0.001$; Table~\ref{tab:headline}), both declining gracefully as the pool grows, while the codebook predecessor never gets off the floor. Right: continuous attention beats the discrete codebook $2$--$10\times$ on every sub-modality.}
\end{figure}

\paragraph{Scaling and the codebook comparison.} Figure~\ref{fig:scaling} traces recall against memory size on the face pool under the paired protocol ($20$ draws per $N$). The training-free read is coincident with the encoder ceiling at every size --- $0.948$ vs.\ $0.948$ at $N{=}10$, $0.777\pm0.008$ vs.\ $0.776\pm0.008$ at $N{=}1000$, $|\Delta|\le0.001$ throughout (Table~\ref{tab:headline}) --- so the memory declines exactly as the encoder's cosine does, gracefully, from near-perfect at ten identities to $0.78$ at a thousand. The coincidence requires the sharp read of Section~\ref{sec:method}: with a soft read (learned inv-temp $\approx 20$) the same sweep trails the cosine by nine to eighteen points at $N\ge300$, as recorded in the paired-protocol paragraph above. The discrete-codebook predecessor (\patha{}), by contrast, flatlines at $\sim$0.07 regardless of codebook size or budget. Figure~\ref{fig:pivot} makes the same point across all five sub-modalities: replacing a discrete codebook with continuous attention lifts recall $2$--$10\times$, because a categorical bottleneck discards the signal the encoder preserved. This is the one place a design choice in the read matters; everything else inherits the encoder.

\paragraph{Vision-language model, key/value orthogonality.} Table~\ref{tab:vlm} gives the numbers behind the structural rule of Section~\ref{sec:when}. Each memory row is compared to retrieval \emph{in its own key space}, under the training-free sharp read. An external ArcFace key (orthogonal to the model's hidden space) reproduces the encoder's recall exactly at every size --- $0.948$/$0.792$/$0.776$ at $N{=}10$/$100$/$1000$, the same values as on the text-LM hosts, as the universality result predicts. The VLM's native vision tokens (already in hidden space) also tie their own retrieval --- the read is faithful to whatever key it is given --- but that ceiling sits far below ArcFace's throughout, so they are the wrong key regardless of method: the empirical basis for reserving the VLM for localization and an external encoder for identity.

\begin{table}[h]
\small\centering
\caption{Parametric memory inside Qwen2.5-VL-3B: same frozen model, only the key encoder changes; training-free sharp read throughout. ArcFace rows: paired protocol, $20$ draws. Native-token rows: single draw on AgeDB with this harness's own key pooling, so they differ from the AgeDB sweep of Section~\ref{sec:when} ($0.64$/$0.34$); both agree that native tokens sit far below ArcFace. The read ties retrieval in both key spaces; only the key's own ceiling differs. An earlier trained variant of the ArcFace row ($12$k steps, learned soft temperature) trailed retrieval at scale ($0.494$ at $N{=}1000$) --- the same under-sharpening artifact recorded in the paired-protocol paragraph above.}
\label{tab:vlm}
\begin{tabular}{@{}lrrr@{}}
\toprule
\textbf{Configuration} & $N{=}10$ & $N{=}100$ & $N{=}1000$ \\
\midrule
Retrieval over ArcFace embeddings (encoder ceiling) & 0.948 & 0.792 & 0.776 \\
\attmem{} $+$ external ArcFace keys (orthogonal to hidden space) & 0.948 & 0.792 & 0.776 \\
\midrule
Retrieval over native vision tokens & 0.50 & 0.43 & --- \\
\attmem{} $+$ native vision tokens (co-located with hidden space) & 0.50 & 0.42 & --- \\
\bottomrule
\end{tabular}
\end{table}

\paragraph{Realistic-scene stress test.} The composites of Section~\ref{sub:factored} place each face in its own well-separated cell, which is where grounding is near-lossless. Table~\ref{tab:realistic} stresses the grounding step instead: faces are pasted onto a dimmed photographic background at jittered positions and varying scales ($0.7$--$1.05\times$), in a one-row or two-dimensional grid layout, and the referent is named by reading order (Qwen2.5-VL-7B, $M{=}40$, $K{=}3$, $3$ seeds per layout). Grounding accuracy drops from $1.00$ on clean composites to $0.81$ (row) and $0.63$ (grid), and end-to-end recall falls well below its oracle because the face detector also degrades on small, rescaled faces (hit-rate $0.09$--$0.43$, forcing un-aligned fallback crops). This is the basis for limitation (iii): in cluttered photographic scenes the grounder and detector, not the encoder, are the practical bottleneck.

\begin{table}[h]
\small\centering
\caption{Realistic-scene stress test: faces composited onto photographic backgrounds with positional jitter and scale variation (Qwen2.5-VL-7B, $M{=}40$, $K{=}3$; mean over $3$ seeds). The correct-region oracle stays high, while VLM grounding and detector alignment degrade --- the failure mode is localization, not identity.}
\label{tab:realistic}
\begin{tabular}{@{}lrrrr@{}}
\toprule
\textbf{Layout} & \textbf{Oracle region} & \textbf{Grounded (ours)} & \textbf{Whole-scene} & \textbf{Grounding acc.} \\
\midrule
Row (jittered)  & 0.83 & 0.42 & 0.03 & 0.81 \\
Grid (jittered) & 0.83 & 0.27 & 0.03 & 0.63 \\
\bottomrule
\end{tabular}
\end{table}

\section{Mechanics of a single recall}
\label{app:anatomy}

It helps to watch one recall end to end. Suppose the agent has met ten people and wants to register an eleventh from a single photo. Registration is one row appended to the face bank (Listing~\ref{lst:row}). The encoder turns the photo into a 512-dimensional ArcFace embedding, the \emph{key}. The model's own embedding for a freshly assigned marker token becomes the \emph{value}, a vector of the model's hidden width (2048 for Qwen2.5-3B). There is no gradient step and no fine-tuning. The row is the memory.

\begin{figure}[h]
\begin{lstlisting}[caption={One registered identity is one row. The key is the encoder's view of the perception; the value is the model's own vector for the marker token that names it. Adding the row is a single tensor append.},label={lst:row}]
# Register identity #11 from a single photo -- no training:
k = l2_normalize(arcface(photo))        # key:   R^512  (encoder space)
m = assign_marker_token()               # e.g. "<id_11>"
v = model.input_embedding[m]            # value: R^2048 (model's own space)
bank.K = torch.cat([bank.K, k[None]])   # O(1) append
bank.V = torch.cat([bank.V, v[None]])
\end{lstlisting}
\end{figure}

At recall, a new photo of one of the eleven arrives under different lighting and a year older. Its ArcFace embedding becomes the query $\bm{q}$. Just before its output head, the model scores $\bm{q}$ against all eleven keys, softmaxes into attention weights $\bm{w}$, and pulls back the weighted blend of values $\bm{r}=\bm{w}^\top\bm{V}$. Because each value is the model's own vector for a marker token, adding $g\,W_o\bm{r}$ to the hidden state lands as a clean boost on the right marker's logit. The model's next token \emph{is} the remembered identity. The whole step is a few matrix multiplies dwarfed by the model's forward pass.

The $0.98$-vs-$0.46$ diagonal gap of Section~\ref{sec:results} (Figure~\ref{fig:mechanism}) is measured on exactly this probe: a held-out ten-identity bank, comparing the sharp attention weights against the raw encoder cosine they are computed from. The attention concentrates the cosine onto its top match, which is why the read recovers the encoder's nearest neighbour as a single decisive marker rather than a diffuse blend.

\section{A discrete-codebook alternative}
\label{app:patha}

Before continuous attention we spent sixteen development cycles on \patha{}: a discrete codebook that snaps each perception to one of $K$ learned codes and remembers the code. It is the natural ``compress the perception into a symbol'' design, and it is a learned cousin of captioning, which is why its failure is instructive rather than embarrassing.

No setting broke it. Across codebook sizes $K\in\{128,256,512,1024\}$, even after \emph{100{,}000} steps of continual pretraining on the expanded face pool, top-1 recall at 300 identities stays pinned between 0.057 and 0.070 while embedding retrieval over the same encoder reaches 0.73. That is a roughly $10\times$ gap that does not move with $K$ (Table~\ref{tab:patha}). Swapping ArcFace R50 for AntelopeV2 R100 trained on 360K identities did not help either.

\begin{table}[h]
\small\centering
\caption{\patha{} after 100K-step continual pretraining on the 2180-identity face pool, at 300 registered identities. Recall is pinned near 0.07 regardless of codebook size, even when the gate routes to the correct code about half the time.}
\label{tab:patha}
\begin{tabular}{@{}lrrrr@{}}
\toprule
\textbf{Codebook size $K$} & 128 & 256 & 512 & 1024 \\
\midrule
\patha{} recall@1            & 0.057 & 0.064 & 0.070 & 0.070 \\
gate routes to correct code  & 0.43  & 0.51  & 0.42  & 0.54  \\
embedding retrieval (same encoder) & \multicolumn{4}{c}{0.73} \\
\bottomrule
\end{tabular}
\end{table}

The reason is a vice the codebook cannot escape, and the diagnostics name it exactly. Two pressures pull in opposite directions as $K$ grows. A \emph{small} codebook packs many people into each cell: at $K{=}16$, distinct identities collide in the same code 8.7\% of the time, and once two people share a code they are indistinguishable forever after. A \emph{large} codebook fixes that, with inter-identity collisions falling to 2.4\% at $K{=}128$, but it shatters each identity across cells. The rate at which two cross-condition photos of the \emph{same} person land in the same code drops from 0.33 to 0.20, so the query no longer routes to where the registration was stored. Net recall is squeezed from both sides and never escapes $\sim$0.07. Even when the gate does route a query to the right code (about half the time, Table~\ref{tab:patha}), every other identity sharing that cell is an equally good answer, so the final pick is near-random.

This is the same information loss as captioning, learned instead of written: any categorical bottleneck, whether a word or a code, throws away the continuous signal the encoder worked to preserve, and no amount of compute downstream can recover it. Continuous attention over the raw embedding never quantizes, and never pays this tax. \patha{} is in the paper because its failure is the cleanest possible argument for the design that replaced it.

\bibliographystyle{plainnat}
\bibliography{refs}

@article{mem0,
  title = {{Mem0}: Building Production-Ready {AI} Agents with Scalable Long-Term Memory},
  author = {Chhikara, Prateek and Khant, Dev and Aryan, Saket and Singh, Taranjeet and Yadav, Deshraj},
  journal = {arXiv preprint arXiv:2504.19413},
  year = {2025}
}

@inproceedings{memllm,
  title = {{MemoryLLM}: Towards Self-Updatable Large Language Models},
  author = {Wang, Yu and Gao, Yifan and Chen, Xiusi and Jiang, Haoming and Li, Shiyang and Yang, Jingfeng and Yin, Qingyu and Li, Zheng and Li, Xian and Yin, Bing and Shang, Jingbo and McAuley, Julian},
  booktitle = {International Conference on Machine Learning (ICML)},
  year = {2024}
}

@article{memgpt,
  title = {{MemGPT}: Towards {LLMs} as Operating Systems},
  author = {Packer, Charles and Wooders, Sarah and Lin, Kevin and Fang, Vivian and Patil, Shishir G and Stoica, Ion and Gonzalez, Joseph E},
  journal = {arXiv preprint arXiv:2310.08560},
  year = {2023}
}

@misc{letta,
  title = {Letta: Stateful {LLM} Agents with Long-Term Memory},
  author = {{Letta Team}},
  year = {2024},
  howpublished = {Software framework},
  url = {https://github.com/letta-ai/letta}
}

@article{m3agent,
  title = {Seeing, Listening, Remembering, and Reasoning: A Multimodal Agent with Long-Term Memory},
  author = {Long, Lin and He, Yichen and Ye, Wentao and Pan, Yiyuan and Lin, Yuan and Li, Hang and Zhao, Junbo and Li, Wei},
  journal = {arXiv preprint arXiv:2508.09736},
  year = {2025}
}

@inproceedings{longmemeval,
  title = {{LongMemEval}: Benchmarking Chat Assistants on Long-Term Interactive Memory},
  author = {Wu, Di and Wang, Hongwei and Yu, Wenhao and Zhang, Yuwei and Chang, Kai-Wei and Yu, Dong},
  booktitle = {International Conference on Learning Representations (ICLR)},
  year = {2025}
}

@inproceedings{myvlm,
  title = {{MyVLM}: Personalizing {VLMs} for User-Specific Queries},
  author = {Alaluf, Yuval and Richardson, Elad and Tulyakov, Sergey and Aberman, Kfir and Cohen-Or, Daniel},
  booktitle = {European Conference on Computer Vision (ECCV)},
  year = {2024}
}

@inproceedings{yollava,
  title = {{Yo'LLaVA}: Your Personalized Language and Vision Assistant},
  author = {Nguyen, Thao and Liu, Haotian and Li, Yuheng and Cai, Mu and Ojha, Utkarsh and Lee, Yong Jae},
  booktitle = {Advances in Neural Information Processing Systems (NeurIPS)},
  year = {2024}
}

@article{mcllava,
  title = {{MC-LLaVA}: Multi-Concept Personalized Vision-Language Model},
  author = {An, Ruichuan and Yang, Sihan and Zhang, Renrui and Lu, Ming and Jiang, Tianyi and Zeng, Kai and Luo, Yulin and Cao, Jiajun and Liang, Hao and Chen, Ying and She, Qi and Zhang, Shanghang and Zhang, Wentao},
  journal = {arXiv preprint arXiv:2411.11706},
  year = {2024}
}

@article{onlinepvlm,
  title = {{Online-PVLM}: Advancing Personalized {VLMs} with Online Concept Learning},
  author = {Bai, Huiyu and Wang, Runze and Du, Zhuoyun and Zhao, Yiyang and Zhang, Fengji and Chen, Haoyu and Zhu, Xiaoyong and Zheng, Bo and Zhao, Xuejiao},
  journal = {arXiv preprint arXiv:2511.20056},
  year = {2025}
}

@inproceedings{rap,
  title = {{RAP}: Retrieval-Augmented Personalization for Multimodal Large Language Models},
  author = {Hao, Haoran and Han, Jiaming and Li, Changsheng and Li, Yu-Feng and Yue, Xiangyu},
  booktitle = {Conference on Computer Vision and Pattern Recognition (CVPR)},
  year = {2025}
}

@inproceedings{tame,
  title = {{TAMEing} Long Contexts in Personalization: Towards Training-Free and State-Aware {MLLM} Personalized Assistant},
  author = {Hong, Rongpei and Lang, Jian and Zhong, Ting and Wang, Yong and Zhou, Fan},
  booktitle = {ACM SIGKDD Conference on Knowledge Discovery and Data Mining (KDD)},
  year = {2026},
  note = {arXiv:2512.21616}
}

@inproceedings{khandelwal2020nearest,
  title = {Generalization through Memorization: Nearest Neighbor Language Models},
  author = {Khandelwal, Urvashi and Levy, Omer and Jurafsky, Dan and Zettlemoyer, Luke and Lewis, Mike},
  booktitle = {International Conference on Learning Representations (ICLR)},
  year = {2020}
}

@inproceedings{wu2022memorizing,
  title = {Memorizing Transformers},
  author = {Wu, Yuhuai and Rabe, Markus N and Hutchins, DeLesley and Szegedy, Christian},
  booktitle = {International Conference on Learning Representations (ICLR)},
  year = {2022}
}

@inproceedings{retro,
  title = {Improving Language Models by Retrieving from Trillions of Tokens},
  author = {Borgeaud, Sebastian and Mensch, Arthur and Hoffmann, Jordan and Cai, Trevor and Rutherford, Eliza and Millican, Katie and van den Driessche, George and others},
  booktitle = {International Conference on Machine Learning (ICML)},
  year = {2022}
}

@inproceedings{flamingo,
  title = {Flamingo: A Visual Language Model for Few-Shot Learning},
  author = {Alayrac, Jean-Baptiste and Donahue, Jeff and Luc, Pauline and Miech, Antoine and Barr, Iain and Hasson, Yana and Lenc, Karel and others},
  booktitle = {Advances in Neural Information Processing Systems (NeurIPS)},
  year = {2022}
}

@inproceedings{lora,
  title = {{LoRA}: Low-Rank Adaptation of Large Language Models},
  author = {Hu, Edward J and Shen, Yelong and Wallis, Phillip and Allen-Zhu, Zeyuan and Li, Yuanzhi and Wang, Shean and Wang, Lu and Chen, Weizhu},
  booktitle = {International Conference on Learning Representations (ICLR)},
  year = {2022}
}

@inproceedings{arcface,
  title = {{ArcFace}: Additive Angular Margin Loss for Deep Face Recognition},
  author = {Deng, Jiankang and Guo, Jia and Xue, Niannan and Zafeiriou, Stefanos},
  booktitle = {Conference on Computer Vision and Pattern Recognition (CVPR)},
  year = {2019}
}

@inproceedings{retinaface,
  title = {{RetinaFace}: Single-Shot Multi-Level Face Localisation in the Wild},
  author = {Deng, Jiankang and Guo, Jia and Ververas, Evangelos and Kotsia, Irene and Zafeiriou, Stefanos},
  booktitle = {Conference on Computer Vision and Pattern Recognition (CVPR)},
  year = {2020}
}

@inproceedings{ecapa,
  title = {{ECAPA-TDNN}: Emphasized Channel Attention, Propagation and Aggregation in {TDNN} Based Speaker Verification},
  author = {Desplanques, Brecht and Thienpondt, Jenthe and Demuynck, Kris},
  booktitle = {INTERSPEECH},
  year = {2020}
}

@inproceedings{wav2vec2,
  title = {{wav2vec~2.0}: A Framework for Self-Supervised Learning of Speech Representations},
  author = {Baevski, Alexei and Zhou, Henry and Mohamed, Abdelrahman and Auli, Michael},
  booktitle = {Advances in Neural Information Processing Systems (NeurIPS)},
  year = {2020}
}

@inproceedings{ast,
  title = {{AST}: Audio Spectrogram Transformer},
  author = {Gong, Yuan and Chung, Yu-An and Glass, James},
  booktitle = {INTERSPEECH},
  year = {2021}
}

@inproceedings{clip,
  title = {Learning Transferable Visual Models From Natural Language Supervision},
  author = {Radford, Alec and Kim, Jong Wook and Hallacy, Chris and Ramesh, Aditya and Goh, Gabriel and Agarwal, Sandhini and Sastry, Girish and Askell, Amanda and Mishkin, Pamela and Clark, Jack and others},
  booktitle = {International Conference on Machine Learning (ICML)},
  year = {2021}
}

@article{dinov2,
  title = {{DINOv2}: Learning Robust Visual Features without Supervision},
  author = {Oquab, Maxime and Darcet, Timoth{\'e}e and Moutakanni, Th{\'e}o and Vo, Huy and Szafraniec, Marc and Khalidov, Vasil and Fernandez, Pierre and others},
  journal = {Transactions on Machine Learning Research (TMLR)},
  year = {2024}
}

@inproceedings{whisper,
  title = {Robust Speech Recognition via Large-Scale Weak Supervision},
  author = {Radford, Alec and Kim, Jong Wook and Xu, Tao and Brockman, Greg and McLeavey, Christine and Sutskever, Ilya},
  booktitle = {International Conference on Machine Learning (ICML)},
  year = {2023}
}

@inproceedings{blip2,
  title = {{BLIP-2}: Bootstrapping Language-Image Pre-training with Frozen Image Encoders and Large Language Models},
  author = {Li, Junnan and Li, Dongxu and Savarese, Silvio and Hoi, Steven},
  booktitle = {International Conference on Machine Learning (ICML)},
  year = {2023}
}

@inproceedings{llava,
  title = {Visual Instruction Tuning},
  author = {Liu, Haotian and Li, Chunyuan and Wu, Qingyang and Lee, Yong Jae},
  booktitle = {Advances in Neural Information Processing Systems (NeurIPS)},
  year = {2023}
}

@inproceedings{sbert,
  title = {Sentence-{BERT}: Sentence Embeddings using {Siamese BERT-Networks}},
  author = {Reimers, Nils and Gurevych, Iryna},
  booktitle = {Empirical Methods in Natural Language Processing (EMNLP)},
  year = {2019}
}

@article{diarizationreview,
  title = {A Review of Speaker Diarization: Recent Advances with Deep Learning},
  author = {Park, Tae Jin and Kanda, Naoyuki and Dimitriadis, Dimitrios and Han, Kyu J and Watanabe, Shinji and Narayanan, Shrikanth},
  journal = {Computer Speech \& Language},
  volume = {72},
  year = {2022}
}

@inproceedings{lfw,
  title = {Labeled Faces in the Wild: A Database for Studying Face Recognition in Unconstrained Environments},
  author = {Huang, Gary B and Mattar, Marwan and Berg, Tamara and Learned-Miller, Eric},
  booktitle = {Workshop on Faces in Real-Life Images},
  year = {2008}
}

@inproceedings{agedb,
  title = {{AgeDB}: The First Manually Collected, In-the-Wild Age Database},
  author = {Moschoglou, Stylianos and Papaioannou, Athanasios and Sagonas, Christos and Deng, Jiankang and Kotsia, Irene and Zafeiriou, Stefanos},
  booktitle = {CVPR Workshops},
  year = {2017}
}

@inproceedings{librispeech,
  title = {{LibriSpeech}: An {ASR} Corpus Based on Public Domain Audio Books},
  author = {Panayotov, Vassil and Chen, Guoguo and Povey, Daniel and Khudanpur, Sanjeev},
  booktitle = {IEEE International Conference on Acoustics, Speech and Signal Processing (ICASSP)},
  year = {2015}
}

@inproceedings{voxceleb,
  title = {{VoxCeleb}: A Large-Scale Speaker Identification Dataset},
  author = {Nagrani, Arsha and Chung, Joon Son and Zisserman, Andrew},
  booktitle = {INTERSPEECH},
  year = {2017}
}

@inproceedings{voxconverse,
  title = {Spot the Conversation: Speaker Diarisation in the Wild},
  author = {Chung, Joon Son and Huh, Jaesung and Nagrani, Arsha and Afouras, Triantafyllos and Zisserman, Andrew},
  booktitle = {INTERSPEECH},
  year = {2020}
}

@inproceedings{ami,
  title = {The {AMI} Meeting Corpus: A Pre-announcement},
  author = {Carletta, Jean and Ashby, Simone and Bourban, Sebastien and Flynn, Mike and Guillemot, Mael and Hain, Thomas and Kadlec, Jaroslav and others},
  booktitle = {International Workshop on Machine Learning for Multimodal Interaction (MLMI)},
  year = {2005}
}

@inproceedings{esc50,
  title = {{ESC}: Dataset for Environmental Sound Classification},
  author = {Piczak, Karol J},
  booktitle = {ACM Multimedia},
  year = {2015}
}

@article{ravdess,
  title = {The {Ryerson} Audio-Visual Database of Emotional Speech and Song ({RAVDESS})},
  author = {Livingstone, Steven R and Russo, Frank A},
  journal = {PLoS ONE},
  volume = {13},
  number = {5},
  pages = {e0196391},
  year = {2018}
}

@misc{wikiart,
  title = {{WikiArt}: Visual Art Encyclopedia},
  author = {{WikiArt}},
  year = {2010},
  howpublished = {\url{https://www.wikiart.org}}
}

@article{qwen25,
  title = {{Qwen2.5} Technical Report},
  author = {Yang, An and Yang, Baosong and Zhang, Beichen and others},
  journal = {arXiv preprint arXiv:2412.15115},
  year = {2024}
}

@article{qwen3,
  title = {{Qwen3} Technical Report},
  author = {Yang, An and Li, Anfeng and Yang, Baosong and others},
  journal = {arXiv preprint arXiv:2505.09388},
  year = {2025}
}

@article{qwen25vl,
  title = {{Qwen2.5-VL} Technical Report},
  author = {Bai, Shuai and Chen, Keqin and Liu, Xuejing and others},
  journal = {arXiv preprint arXiv:2502.13923},
  year = {2025}
}

@article{qwen25omni,
  title = {{Qwen2.5-Omni} Technical Report},
  author = {Xu, Jin and Guo, Zhifang and He, Jinzheng and others},
  journal = {arXiv preprint arXiv:2503.20215},
  year = {2025}
}

@article{phi3,
  title = {{Phi-3} Technical Report: A Highly Capable Language Model Locally on Your Phone},
  author = {Abdin, Marah and Jacobs, Sam Ade and Awan, Ammar Ahmad and others},
  journal = {arXiv preprint arXiv:2404.14219},
  year = {2024}
}

@article{smollm2,
  title = {{SmolLM2}: When Smol Goes Big --- Data-Centric Training of a Small Language Model},
  author = {Ben Allal, Loubna and Lozhkov, Anton and Bakouch, Elie and others},
  journal = {arXiv preprint arXiv:2502.02737},
  year = {2025}
}

@article{mistral7b,
  title = {Mistral {7B}},
  author = {Jiang, Albert Q and Sablayrolles, Alexandre and Mensch, Arthur and others},
  journal = {arXiv preprint arXiv:2310.06825},
  year = {2023}
}

@article{deepseekr1,
  title = {{DeepSeek-R1}: Incentivizing Reasoning Capability in {LLMs} via Reinforcement Learning},
  author = {{DeepSeek-AI}},
  journal = {arXiv preprint arXiv:2501.12948},
  year = {2025}
}

@misc{granite4,
  title = {Granite 4.0: Hyper-Efficient, High-Performance Hybrid Models for Enterprise},
  author = {{IBM Granite Team}},
  year = {2025},
  howpublished = {IBM announcement},
  url = {https://www.ibm.com/new/announcements/ibm-granite-4-0-hyper-efficient-high-performance-hybrid-models}
}

@article{userengram,
  title = {User as Engram: Internalizing Per-User Memory as Local Parametric Edits},
  author = {Li, Bojie},
  journal = {arXiv preprint arXiv:2606.19172},
  year = {2026}
}

@article{usercode,
  title = {User as Code: Executable Memory for Personalized Agents},
  author = {Li, Bojie},
  journal = {arXiv preprint arXiv:2606.16707},
  year = {2026}
}

@article{editablekv,
  title = {Models Take Notes at Prefill: {KV} Cache Can Be Editable and Composable},
  author = {Li, Bojie},
  journal = {arXiv preprint arXiv:2606.17107},
  year = {2026}
}

@article{latentbridge,
  title = {The Latent Bridge: A Continuous Slow--Fast Channel for Real-Time Game Agents},
  author = {Li, Bojie and Shi, Noah},
  journal = {arXiv preprint arXiv:2606.24470},
  year = {2026}
}

@misc{pineai2026whispercoding,
  title        = {{Pine AI}: The Most Natural Human-Computer Interface Is Your Voice},
  author       = {{Pine AI}},
  year         = {2026},
  howpublished = {Blog post},
  url          = {https://www.19pine.ai/blog/pine-ai-the-most-natural-human-computer-interface-is-your-voice},
  note         = {Accessed 2026-06-28}
}

\end{document}